\documentclass[10pt,twocolumn,letterpaper]{article}
\usepackage{multirow}
\usepackage{subcaption} 
\usepackage{bm, float, caption, subcaption, graphicx}
\usepackage{amssymb}
\usepackage{array}
\usepackage{xcolor}
\usepackage{marvosym}
\definecolor{greenbg}{rgb}{0.9, 1.0, 0.9}
\definecolor{lowred}{RGB}{238,18,137}
\definecolor{lowerred}{RGB}{255,110,180}
\definecolor{rowunit}{RGB}{128,128,255}
\definecolor{evaunit01green}{RGB}{54,125,189}
\definecolor{cvprblue}{rgb}{0.21,0.49,0.74}

\newcommand{\evagreen}[1]{\textcolor{evaunit01green}{#1}}
\newcommand{\dplus}[1]{\fontsize{6pt}{0.1em}\selectfont (\textbf{\textcolor{lowred}{#1}})}
\newcommand{\dtplus}[1]{\fontsize{6pt}{0.1em}\selectfont (\textbf{\evagreen{#1}})}
\newcommand{\supptitle}[1]{
        \hspace{-15pt}
        {\Large\bfseries #1\par}
        \vspace*{10pt}
}

\usepackage[pagenumbers]{cvpr} % To force page numbers, e.g. for an arXiv version       
\usepackage[pagebackref,breaklinks,colorlinks,allcolors=cvprblue]{hyperref}
\usepackage{xcolor}
\usepackage[table]{xcolor}
\usepackage{wrapfig}  

\def\paperID{5913} 
\def\confName{CVPR}
\def\confYear{2026}

\title{PointTPA: Dynamic Network Parameter Adaptation for \\3D Scene Understanding}

\author{Siyuan Liu$^{*}$, Chaoqun Zheng$^{*}$, Xin Zhou, Tianrui Feng, Dingkang Liang$^{\text{\Letter}}$, Xiang Bai\\
Huazhong University of Science and Technology \\ {\tt\{syliu,cqzheng,dkliang\}@hust.edu.cn}}

\begin{document}

\maketitle
{\let\thefootnote\relax\footnotetext{\hspace{-1.5em}* Equal contribution. $\text{\Letter}$ Corresponding author (dkliang@hust.edu.cn).}}
 
\begin{abstract}

\noindent Scene-level point cloud understanding remains challenging due to diverse geometries, imbalanced category distributions, and highly varied spatial layouts. Existing methods improve object-level performance but rely on static network parameters during inference, limiting their adaptability to dynamic scene data. We propose PointTPA, a \textbf{T}est-time \textbf{P}arameter \textbf{A}daptation framework that generates input-aware network parameters for scene-level point clouds. PointTPA adopts a Serialization-based Neighborhood Grouping (SNG) to form locally coherent patches and a Dynamic Parameter Projector (DPP) to produce patch-wise adaptive weights, enabling the backbone to adjust its behavior according to scene-specific variations while maintaining a low parameter overhead. Integrated into the PTv3 structure, PointTPA demonstrates strong parameter efficiency by introducing two lightweight modules of less than 2\% of the backbone's parameters. Despite this minimal parameter overhead, PointTPA achieves 78.4\% mIoU on ScanNet validation, surpassing existing parameter-efficient fine-tuning (PEFT) methods across multiple benchmarks, highlighting the efficacy of our test-time dynamic network parameter adaptation mechanism in enhancing 3D scene understanding. The code is available at \url{https://github.com/H-EmbodVis/PointTPA}.

\end{abstract}   
\section{Introduction}
\label{sec:intro}

\begin{figure*}[h]
  \centering
  \includegraphics[width=1\textwidth]{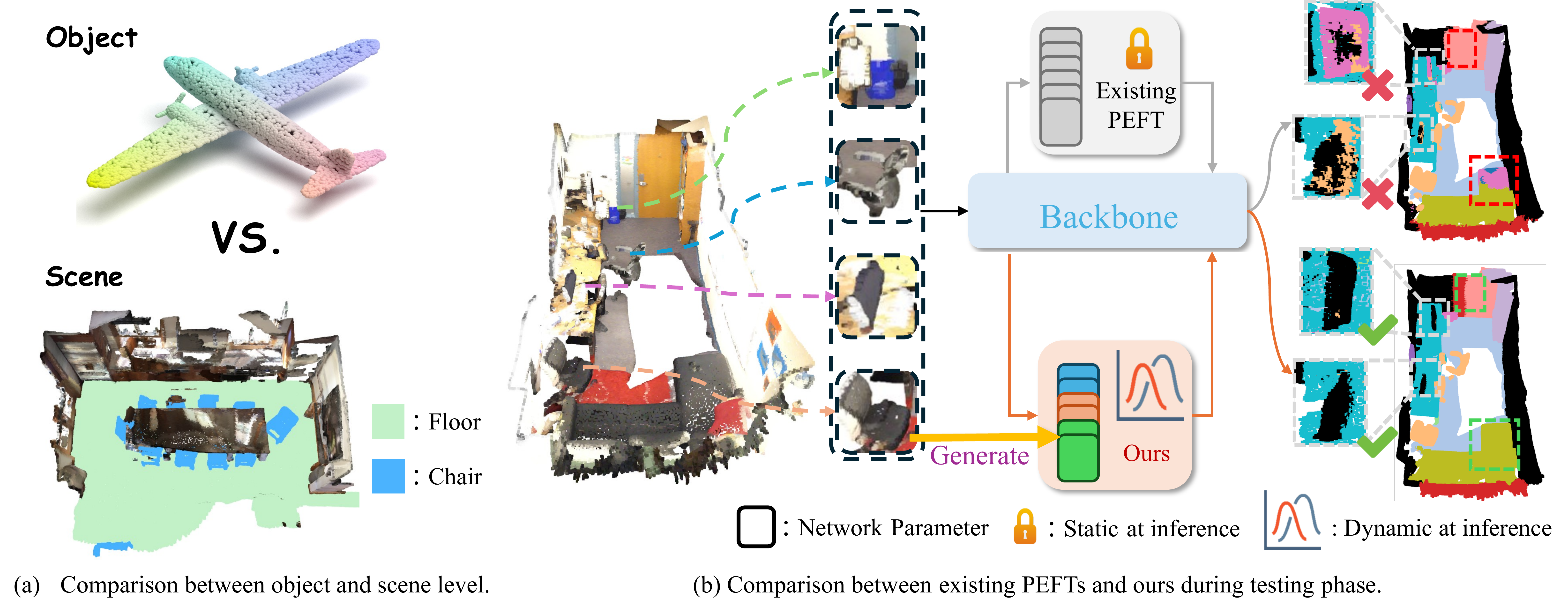}
  \vspace{-10pt}
  \caption{(a) Scene-level point clouds have more points and highly imbalanced category distributions compared to object-level point clouds. (b) Different from the existing PEFT methods, our PointTPA provides dynamic projection weights during inference and performs better in 3D semantic segmentation.}
  \label{fig:intro}
\end{figure*}

3D scene perception via point clouds plays an important role in embodied intelligence and autonomous driving~\cite{guo2020deep, lu2022transformers,fan2023super, li2026sgdrive,liang2026cook,liang2025seeing}. By directly sampling geometric information, point clouds enable precise modeling of the 3D world. However, their irregular and sparse structures pose distinct challenges for effective processing.

Recent point cloud perception research~\cite{zhao2021point,guo2021pct,lu2022transformers,yu2022point, zha2023sfr, zhang2023simple} significantly advances 3D scene understanding by leveraging transformer architectures, which excel at modeling long-range dependencies in point sets, demonstrating high scalability, and offering exceptional transferability to diverse downstream tasks. Moreover, large-scale self-supervised pre-training approaches~\cite{xie2020pointcontrast,wu2023masked,wu2025sonata,jiang2020pointgroup} further strengthen the representational capacity of these foundation models. Among these, Sonata~\cite{wu2025sonata} exhibits remarkable capabilities by alleviating local geometric shortcut problems through random position disturbing and self-distillation.

Despite the strong inductive biases and rich prior knowledge encoded by pre-training, foundation models often fail to fully realize their potential when fine-tuned on downstream tasks for scene-level point clouds. This performance gap stems from intrinsic challenges in 3D scenes, which hinder effective knowledge transfer and adaptation: 1) Diverse local geometries: Scene-level point clouds contain many more points compared with object-level point clouds. Scene-level point clouds are distributed across multiple objects and background regions, forming intricate and dynamic geometric patterns. 2) Imbalanced category distributions: Scene-level point clouds encompass a wide range of categories, often with highly imbalanced point distributions. As shown in Fig.~\ref{fig:intro}, common categories (\eg, floor) dominate the point cloud in terms of point count, whereas rare objects, such as small objects (\eg, chairs), occupy only a tiny fraction of points, resulting in severe inter-class imbalance. 3) Variable spatial organizations: At the scene level, points are spread across expansive and heterogeneous spatial layouts, with objects varying in position, orientation, and scale depending on the scene dynamics. These challenges highlight the importance of a model's ability to dynamically adapt to ensure robust and context-aware scene-level point cloud understanding.

To mitigate these limitations, researchers have explored input-adaptive and dynamic fine-tuning approaches that adjust model behavior based on the input while preserving pretrained knowledge of frozen 3D foundation models. Methods derived from parameter-efficient fine-tuning (PEFT), such as IDPT~\cite{zha2023instance}, DAPT~\cite{zhou2024dynamic}, and PointGST~\cite{liang2025parameter}, introduce dynamic object-aware mechanisms to enhance performance. However, these approaches primarily emphasize object-level dynamic adaptation and fall short of addressing the core challenges of large-scale 3D scene understanding, where substantial geometric variation and structural complexity require more powerful and globally consistent input-driven network parameter dynamics.

We propose PointTPA, a framework designed for scene-level point clouds with a dynamic network parameter adaptation mechanism. Unlike existing methods, where the parameters remain static during inference, PointTPA dynamically generates projection weights based on input tokens, enabling the model to adapt its weights to the specific characteristics of each scene during inference. Notably, we build our dynamic inference modules via the PEFT mechanism, which enables lightweight and effective adaptation while preserving the rich semantic and geometric priors learned during pre-training. These designs make the model more sensitive to geometric and semantic variations. In this way, the model maintains parameter efficiency while improving its generalization ability across diverse scenarios.

Extensive experiments on multiple datasets demonstrate the efficiency and robustness of our approach. When integrated into Point Transformer V3 (PTv3)~\cite{wu2024point}, our PointTPA reduces the number of trainable parameters by more than 98\% compared with full fine-tuning of the backbone, while maintaining comparable or even better performance across benchmarks. On the ScanNet validation~\cite{dai2017scannet}, our method achieves 78.4\% mIoU, surpassing existing object-level PEFT methods. Moreover, our approach delivers superior performance on a similar parameter scale. On the ScanNet++~\cite{yeshwanth2023scannet++} dataset, it further outperforms the PointGST~\cite{liang2025parameter} method by 0.9\%, highlighting its strong adaptability to complex scene-level point clouds.

In general, our main contributions can be summarized as follows: \textbf{1)} By revisiting key challenges in 3D scene perception, we highlight that inference-time parameter dynamism is crucial for effectively adapting pretrained models to complex scene-level variations. \textbf{2)} We propose PointTPA, a test-time parameter-dynamic method with low-cost training acquisition for 3D scene perception, enabling the model to adapt to diverse scene-level point cloud variations. \textbf{3)} Through extensive experiments, we demonstrate that PointTPA achieves state-of-the-art performance among PEFT methods and highly competitive results to full fine-tuning across diverse 3D scene perception benchmarks.

\section{Related Work}

\subsection{Scene Perception for Point Cloud}
As a mature representation of the 3D geometry, point clouds play a critical role in scene representation and perception tasks~\cite{qi2017pointnet, qi2017pointnet++, liu2020closer, xiang2021walk, liang2024pointmamba, li2026geoteacher}. Earlier object-centric studies mainly focus on isolated shapes with relatively simple geometric layouts~\cite{zhang2022point, Zhang_2023_CVPR, zha2024towards, zha2024lcm, li2023dds3d}. Driven by advan ces in embodied artificial intelligence and autonomous driving~\cite{li2026imagidrive, zhou2025hermes,fu2025orion,fu2025minddrive, liang2025sood++}, the research focus increasingly shifts toward large-scale scene-level understanding~\cite{xu2024unified}.

Large-scale 3D scene perception is inherently more challenging than object-level analysis due to the diverse local geometries, imbalanced category distributions, and varied spatial organizations. To alleviate these difficulties, point-based architectures process raw points directly using hierarchical sampling and neighborhood aggregation~\cite{qian2022pointnext, guo2021pct, wu2022point}. Voxel-based approaches convert point clouds into regular grids to enable efficient sparse or dense 3D convolutions~\cite{zhou2018voxelnet, mao2021voxel, chen2023voxelnext}. Projection and multimodal alignment methods map points into multi-view or image-conditioned spaces to exploit 2D foundation models~\cite{su2015multi, chen2023clip2scene, zhang2022pointclip, zhu2023pointclip}. Recently, transformer-based models~\cite{zhao2021point, park2023self, wu2023masked, wu2024point, zha2025point} have enhanced scene perception by modeling long-range dependencies and flexible global context aggregation.

\subsection{Self-Supervised Learning for Point Cloud}
To further strengthen the model's generalization ability, self-supervised pre-training methods are widely adopted. In general, self-supervised pretraining for point clouds encompasses two main paradigms. The contrastive learning paradigm~\cite{afham2022crosspoint,chen2023clip2scene,xie2020pointcontrast,hou2021exploring} aims to align representations from different views, while reconstruction-based methods~\cite{pang2022masked,yu2022point,zhang2022point, zha2024pre} learn discriminative features by recovering masked regions of the input. Sonata~\cite{wu2025sonata} introduces a self-distillation pretraining strategy in which the spatial coordinates of masked points are deliberately perturbed during training. This design encourages transformer-based architectures to develop stronger geometric inductive biases, thereby improving their ability to capture spatial structures. Thus, applying these models to various downstream tasks has attracted considerable attention.

Although full fine-tuning (FFT) of the pretrained model is feasible, as pretrained Transformers grow and downstream tasks multiply, computing and memory demands increase sharply. This creates practical constraints for large-scale training and deployment.

\subsection{PEFT Method for Point Cloud Analysis}
To reduce the high costs associated with FFT, PEFT methods~\cite{fei2025parameter,hu2021lora,li2021prefix,valipour2023dylora} emerge as a promising alternative, enabling efficient adaptation by updating only a small fraction of parameters. Inspired by the success of PEFT in 2D vision, researchers have begun to explore its application in point cloud analysis~\cite{zha2023instance,tang2024any2point,zhou2024dynamic,liang2025parameter,sun2024parameter,tang2024point, zha2025pma}. For example, IDPT~\cite{zha2023instance} introduces DGCNN~\cite{wang2019dynamic} into learnable modules to generate instance-aware prompts for pretrained models. Point-PEFT~\cite{tang2024point} aims to capture and leverage local point features during the fine-tuning. DAPT~\cite{zhou2024dynamic} proposes a dynamic scale adapter and integrates it with prompt tuning~\cite{jia2022visual}. PointGST~\cite{liang2025parameter} employs a spectral-domain adapter to integrate features from multiple representations. These approaches achieve impressive progress in object-level point cloud analysis. 

However, these methods exhibit inherent limitations when directly applied to scene-level understanding tasks. Unlike object-level point clouds, scene-level data present additional challenges, such as highly variable geometric structures, imbalanced category distributions, and diverse spatial layouts. These distinctions hinder the adaptability of the above methods, resulting in performance degradation on complex scene-level tasks. To alleviate these issues, we propose PointTPA, which enables more efficient and effective 3D scene understanding with a dynamic mechanism.

\section{Method}
\begin{figure*}[h]
  \centering
  \includegraphics[width=0.99\textwidth]{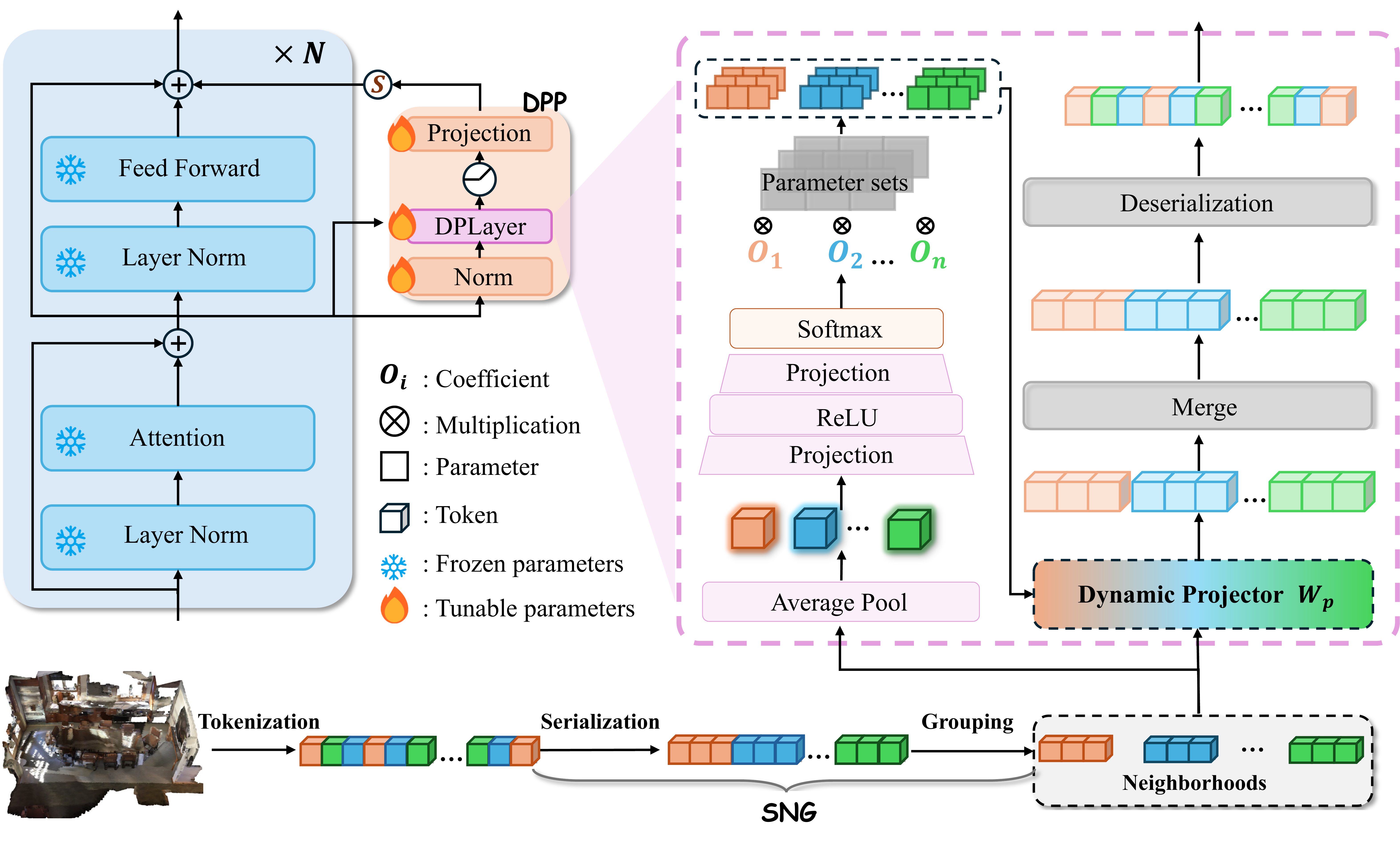}
  \vspace{-5pt}
  \caption{Overview of PointTPA. It consists of a Serialization-based Neighborhood Grouping (SNG) and a Dynamic Parameter Projector (DPP). During the training process, we freeze the backbone parameter and only fine-tune our DPP and static Adapter modules. The DPP can produce dynamic weights during inference. The same color denotes tokens with spatially close positions.}
  \label{fig:pipeline}
  \vspace{-10pt}
\end{figure*}

As shown in Fig.~\ref{fig:pipeline}, PointTPA consists of a Serialization-based Neighborhood Grouping (SNG) and a Dynamic Parameter Projector (DPP), working together in two main stages: 1) Training for Weights Adaptation Modules and 2) Input-Aware Dynamic Parameter Inference. This unified framework facilitates efficient parameter adaptation, enhancing performance across input scenes and datasets while ensuring computational efficiency. A detailed description of the framework is in the subsequent sections.

\subsection{Modules for Weights Adaptation}
During the training process, we freeze the backbone model and fine-tune the modules of PointTPA, which run in parallel with the FFN. The SNG organizes the scene into locally correlated groups, ensuring both geometric and contextual consistency. The DPP learns to generate group-specific projection weights conditioned on local features, enabling adaptive parameter adjustments.

\paragraph{Serialization-based Neighborhood Grouping (SNG).} To efficiently process large-scale 3D scenes, we decompose the naturally unordered point clouds into structured local patches. We achieve this by serializing the input points into a regular sequence with space-filling curves, which preserves space locality and groups neighboring tokens. This organization sets the stage for the subsequent dynamic parameter adaptation. Specifically, we employ established spatial traversal patterns, such as the Hilbert or Z-order curves, to reorder the point cloud data. The serialization operation is formally expressed as:
{
\begin{equation}
\bm{x}_{seq} = \varphi(\operatorname{LN}(\bm{x})),
\label{eq:serialization}
\end{equation}
}
where $\bm{x} \in \mathbb{R}^{N \times C}$ represents the unordered point cloud token, consisting of $N$ points with $C$ features. $\operatorname{LN}$ denotes Layer Normalization. $\varphi$ and $\varphi^{-1}$ denote the serialization and deserialization operations.

Once serialized, we proceed to group the data into local neighborhoods. Given serialized tokens $\bm{x}_{seq}$, we partition this sequence into local groups as follows:
{
\begin{equation}
\bm{x}_g = \mathcal{G}_m(\bm{x}_{seq}),
\label{eq:group}
\end{equation}
}
where $\mathcal{G}_m(\cdot)$ divides the serialized sequence into $m$ groups. Each grouped point cloud feature in $\bm{x}_g \in \mathbb{R}^{m \times C \times n}$ contains up to $n$ points per group. To ensure uniformity, zero-padding is applied to groups with fewer points.

\paragraph{Dynamic Parameter Projector (DPP).}
\label{subsec:DPP} Given the grouped tokens from SNG, our DPP module dynamically generates weights tailored to these tokens. This enables adaptive processing for each input group. The core of the DPP module consists of the dynamic parameter layer (DPLayer) and the parameter base set. The DPLayer employs a dynamic weight router to learn dynamic coefficients for each grouped token during training. These coefficients then interact with the parameter base set, a collection of randomly initialized learnable weight bases, to generate adaptive projection weights. This design enables efficient, context-specific adjustments, allowing the model to dynamically adapt to varying scene-level point clouds.

The parameter base set $\bm{P}\in \mathbb{R}^{K\times C\times C_d}$ consists of a set of learnable weight parameters that are continuously updated during the training process:
\begin{equation}
    \bm{P}=[\bm{P}_{1}, \bm{P}_{2}, \dots, \bm{P}_{K}], 
  \label{eq:co}
\end{equation}
where $K$ is the number of bases of dynamic parameters, $C_d$ is the hidden dimension. Its primary purpose is to capture a rich and adaptable parameter space capable of supporting dynamic adjustments across varying depths of the model. Empirically, a larger $K$ improves dynamic expressivity but increases the number of parameters and training cost. Further discussion can be found in Sec.~\ref{ablation}.

Receiving the grouped tokens $\bm{x}_{g}$ from SNG, the weight router aggregates them into $\bm{x}_p \in \mathbb{R}^{m\times C}$ via average pooling to extract a unified feature descriptor for each patch, and produces the routing coefficients $\bm{O}\in \mathbb{R}^{m\times K}$. Each row of $\bm{O}$ represents the softmax-normalized coefficients on the $K$ bases for a token group. The $i$-th row $\bm{O}^{(i)}$ corresponds to $\bm{x}_g^{(i)}$ ($ i=1,\dots, m$) to generate a specific projection weight:
\begin{equation}
  \begin{split}
    &\bm{x}_p= \operatorname{AvgPool}(\bm{x}_g),\\
    \bm{O} = &\operatorname{Softmax}(\operatorname{MLP}(\bm{x}_p)/ \tau),
  \end{split}
  \label{eq:router}
\end{equation}
where $\operatorname{MLP}$ is a lightweight network consisting of two linear layers with a non-linear activation in between, and $\tau$ refers to the softmax temperature, typically set to 4.

With the parameter base set $\bm P$ and the routing coefficients $\bm{O}$, we can obtain a merged dynamic projection weight $\bm{W}_p\in \mathbb{R}^{m\times C\times C_d}$ for ${\bm{x}}_g$, formulated as:
\begin{equation}
  \begin{split}
    \bm{W}_{p}^{(i)} &= \sum_{k=1}^K \bm{O}_{i,k}\bm{P}_{k}.
  \end{split}
  \label{eq:dw}
\end{equation}
After obtaining the dynamic weights, we employ them as projectors on $\bm{x}_g$ to produce adapted tokens $\bm{x}_{g}'\in \mathbb{R}^{m \times C_d\times n}$. To generate the final output $\bm{x}_\mathrm{out}$, the tokens are first restored to their original order via inverse grouping $\mathcal{G}^{-1}$ and inverse serialization $\varphi^{-1}$. Following a linear projection to match the dimension of $\bm{x}$, these refined features are scaled and added to the initial input:
\begin{equation}
  \begin{split}
    &\bm{x}'_{g} = \left\{\bm{W}_p^{(i)\top} \bm{x}_{g}^{(i)}\right\}_{i=1}^{m},\\
    \bm{x}_\mathrm{out} = \bm{x}&+s\cdot\phi \left [\operatorname{ReLU} \left(\varphi^{-1}\mathcal G^{-1} \left(\bm{x}'_g \right)\right)\right], 
  \end{split}
  \label{eq:DPL}
\end{equation}
where $s$ denotes the scaling factor, while $\phi$ acts as a linear projector from dimension $C_d$ to $C$. $\bm{x}_\mathrm{out}$ is the output of DPP. It aggregates the input-aware features and enables the model to facilitate a more effective extraction of geometric features from complex scenes.

\begin{figure}[t]
\centering
\includegraphics[width=0.99\linewidth]{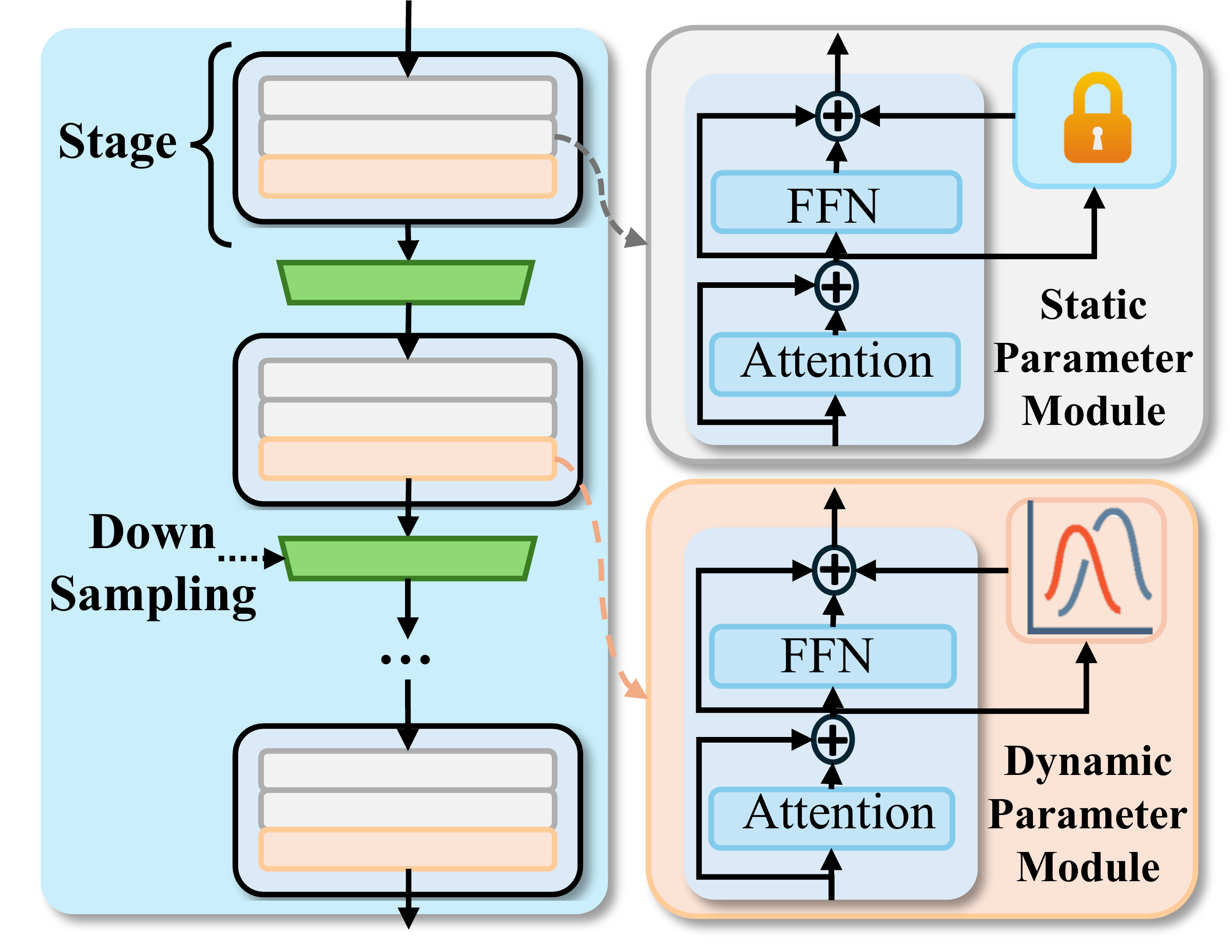}
\vspace{-5pt}
\caption{Illustration of our mixed-insertion strategy. PointTPA is applied to the last block of each stage, while static adapters are adopted in the remaining blocks.}
\label{insertion}
\end{figure}

\paragraph{Mixed-Insertion Strategy.} 
As shown in Fig.~\ref{insertion}, we adopt a mixed-insertion strategy, where the SNG and DPP modules are inserted only into the last block of each encoder stage, while static adapters are retained in the preceding blocks. The static adapters remain fixed in inference to ensure stable and consistent feature processing. In contrast, the DPP module dynamically generates input-dependent weights in both training and inference, enabling flexible adaptation to unseen geometric variations.

\subsection{Input-Aware Dynamic Parameter Inference}

PointTPA achieves dynamic inference by coupling the SNG and DPP modules, allowing the model to adjust its parameters according to the features of the scene-level point cloud.

During inference, the parameter-free SNG module first restructures the input point cloud by grouping points into spatially coherent local patches. This organization preserves geometric locality while providing a structured representation for subsequent processing. The grouped patches are then fed into the DPP module, which generates adaptive parameters conditioned on their content. Specifically, the DPP employs a set of fine-tuned parameter bases together with a coefficient router. For each patch, the DPLayer predicts projection coefficients that linearly combine these bases to produce patch-specific dynamic weights.

Through this mechanism, PointTPA assigns tailored parameters to different regions of the scene, enabling the frozen backbone to better capture local geometric variations. This leads to more robust and efficient feature extraction across diverse and complex scene-level point clouds.

\begin{table*}
  \footnotesize
  \caption{Results on ScanNet Validation~\cite{dai2017scannet}, S3DIS~\cite{armeni20163d}, and ScanNet++~\cite{yeshwanth2023scannet++}.
Only trainable parameters are reported. The \text{dec.} and \text{w/o dec.} indicate Sonata with and without the decoder, respectively. We use linear probing as the baseline and Sonata (w/o dec.) as a strong comparison. The Avg. Rank is the average of all the metrics of these benchmarks.}
  \label{tab:main_result}
  \centering
  \resizebox{\textwidth}{!}{%
  \begin{tabular}{@{}l*{14}{c}@{}}
    \toprule
    
    \multirow{2.3}{*}{ } & \multirow{2.3}{*}{Reference} &  \multicolumn{2}{c}{\multirow{2.3}{*}{Param. (M)}}& \multicolumn{3}{c}{ScanNet Val~\cite{dai2017scannet}} & \multicolumn{3}{c}{S3DIS Area5~\cite{armeni20163d}} & \multicolumn{3}{c}{ScanNet++ Val~\cite{yeshwanth2023scannet++}} & \multirow{2.3}{*}{Avg. Rank}\\
     \cmidrule(lr){5-7} \cmidrule(lr){8-10} \cmidrule(lr){11-13} 
    & & & &mAcc &allAcc&mIoU  &mAcc &allAcc&mIoU &mAcc &allAcc&mIoU & \\
    \midrule
    \multicolumn{14}{l}{\textit{Training from scratch}}\\
    \midrule
    SparseUNet~\cite{choy20194d} &CVPR 19 &\multicolumn{2}{c}{39.2} &80.2&90.0&72.3&72.5&89.8&66.3&38.4&80.1&28.8 & - \\
    PointNeXt-XL~\cite{qian2022pointnext}  &NeurIPS 22   &\multicolumn{2}{c}{41.6}&-&-&71.5&-&90.6&70.5&-&-&- & - \\
    PTv3~\cite{wu2024point} &CVPR 24 &\multicolumn{2}{c}{124.8} &85.0&92.0&77.6&78.9&91.7&73.4&53.4&85.6&42.1 & - \\   
    \midrule
    \multicolumn{14}{l}{\textit{Full Fine-tuning}}\\
    \midrule
     PTv3-PPT (\text{sup.})~\cite{wu2024towards} &CVPR 24  &\multicolumn{2}{c}{124.8} &85.9 &92.3 &78.6 &80.1 &92.0 &74.3 &55.7 &86.4 &43.3  & - \\
     Sonata (\text{dec.})~\cite{wu2025sonata} & CVPR 25 &\multicolumn{2}{c}{124.8} &86.1 &92.5 &79.4 &81.6 &93.0 &76.0 &55.8 &86.6 &43.7 & - \\
    Sonata (\text{w/o dec.})~\cite{wu2025sonata} & CVPR 25 & \multicolumn{2}{c}{108.5} &86.1 &92.3  &78.9 &80.3 &93.1 &74.5  &50.7 &86.3 &41.8 & - \\
    \midrule
    \multicolumn{14}{l}{\textit{General PEFT methods}}\\
    \midrule
    Sonata (linear probing) & &\multicolumn{2}{c}{0.02 (0.02\%)}  & 83.1 & 89.7 & 72.2&80.9 &90.8 &73.0  &49.4 &84.9 &36.5 & - \\
    + Adapter~\cite{houlsby2019parameter}&ICML 19 & \multicolumn{2}{c}{1.90 (1.76\%)} &85.1\dplus{+2.0} &91.7\dplus{+2.0} &76.8\dplus{+4.6}&81.7\dplus{+0.8} &91.6\dplus{+0.8} &73.8\dplus{+0.8} &\textbf{52.9}\dplus{+3.5} &\textbf{86.1}\dplus{+1.2} &39.9\dplus{+3.4} & 3.6 \\
    + Prefix Tuning~\cite{li2021prefix} &ACL 21&\multicolumn{2}{c}{0.08 (0.08\%)}  &84.2\dplus{+1.1}&90.3\dplus{+0.6} &73.5\dplus{+1.3}&\textbf{82.1}\dplus{+1.2} &91.0\dplus{+0.2} & 73.0\dtplus{-0.0} &49.6\dplus{+0.2} &84.2\dtplus{-0.7} &36.8\dplus{+0.3} & 7.3 \\
    + BitFit~\cite{zaken2022bitfit} &ACL 22 &\multicolumn{2}{c}{0.12 (0.11\%)}  &85.4\dplus{+2.3} &91.1\dplus{+1.4} & 75.4\dplus{+3.2}& 81.2\dplus{+0.3}&92.2\dplus{+1.4} &73.8\dplus{+0.8}  &50.7\dplus{+1.3} &84.5\dtplus{-0.4} &38.0\dplus{+1.5} & 5.6 \\
    + LoRA~\cite{hu2021lora}&ICLR 22 &\multicolumn{2}{c}{0.89 (0.82\%)}  &85.2\dplus{+2.1} &91.6\dplus{+1.9} & 76.3\dplus{+4.1}& 81.1\dplus{+0.2}&92.5\dplus{+1.7} &74.0\dplus{+1.0} &51.4\dplus{+2.0} &85.3\dplus{+0.4} &38.7\dplus{+2.2} & 5.0 \\
    + VeRA~\cite{kopiczkovera}&ICLR 24 & \multicolumn{2}{c}{0.08 (0.08\%)} &84.0\dplus{+0.9} &90.2\dplus{+0.5}  &73.4\dplus{+1.2} &80.8\dtplus{-0.1}  &91.0\dplus{+0.2}  &72.9\dtplus{-0.1}  &49.9\dplus{+0.5}  &83.9\dtplus{-1.0}  &36.9\dplus{+0.4} & 9.0 \\
    + RandLoRA~\cite{albert2024randlora}&ICLR 25 &\multicolumn{2}{c}{0.71 (0.66\%)} &84.3\dplus{+1.2} &90.7\dplus{+1.0} &74.3\dplus{+2.1} &81.7\dplus{+0.8}  &91.2\dplus{+0.4}  &73.0\dtplus{-0.0}  &50.1\dplus{+0.7}  &84.1\dtplus{-0.8}  &37.2\dplus{+0.7} & 6.6 \\
    \midrule
    \multicolumn{14}{l}{\textit{PEFT methods for point cloud}}\\
    \midrule
    + IDPT~\cite{zha2023instance} & ICCV 23&\multicolumn{2}{c}{2.61 (2.42\%)}  &82.9\dtplus{-0.2} &89.8\dplus{+0.1} &72.6\dplus{+0.4} &81.2\dplus{+0.3} &90.5\dtplus{-0.3} &72.0\dtplus{-1.0}  &50.0\dplus{+0.6} &84.1\dtplus{-0.8} &36.6\dplus{+0.1} & 9.1 \\
    + DAPT~\cite{zhou2024dynamic} &CVPR 24&\multicolumn{2}{c}{1.14 (1.06\%)} &\textbf{86.5}\dplus{+3.4} &91.9\dplus{+2.2} &77.7\dplus{+5.5}  &81.2\dplus{+0.3} &92.7\dplus{+1.9} &74.6\dplus{+1.6} &50.9\dplus{+1.5} &85.2\dplus{+0.3} &39.6\dplus{+3.1} & 3.4 \\  
    + PointGST~\cite{liang2025parameter}&TPAMI 25 & \multicolumn{2}{c}{1.05 (0.97\%)} &85.8\dplus{+2.7} &91.9\dplus{+2.2} &77.7\dplus{+5.5} &81.4\dplus{+0.5} &\textbf{92.9}\dplus{+2.1} &\textbf{75.0}\dplus{+2.0} &52.0\dplus{+2.6} &85.9\dplus{+1.0} &40.0\dplus{+3.5} & 2.4 \\
    + \textbf{PointTPA (\text{ours})}&-- &\multicolumn{2}{c}{1.18 (1.09\%)}  &86.3\dplus{+3.2} &\textbf{92.3}\dplus{+2.6} &\textbf{78.4}\dplus{+6.2}&81.7\dplus{+0.8} &\textbf{92.9}\dplus{+2.1} &74.9\dplus{+1.9}  &\textbf{52.9}\dplus{+3.5} &\textbf{86.1}\dplus{+1.2} &\textbf{40.9}\dplus{+4.4} & \textbf{1.3} \\
    \bottomrule
  \end{tabular}
  }
  \vspace{-5pt}
\end{table*}

\section{Experiments}
\subsection{Experimental Setup}
To evaluate scene-level semantic segmentation, we benchmark our method on ScanNet~\cite{dai2017scannet}, ScanNet++~\cite{yeshwanth2023scannet++}, and S3DIS~\cite{armeni20163d} using the PTv3~\cite{wu2024point} equipped with pre-trained Sonata~\cite{wu2025sonata,pointcept2023} weights. We use linear probing as our standard baseline. To establish a more rigorous performance bound, we benchmark against the decoder-free full fine-tuning method (referred to as FFT throughout this section).

To demonstrate the superior performance of the proposed approach, we compare our PointTPA with both general PEFT methods~\cite{zaken2022bitfit, hu2021lora, li2021prefix, houlsby2019parameter} and PEFT methods for point cloud analysis, such as IDPT~\cite{zha2023instance}, DAPT~\cite{zhou2024dynamic}, and PointGST~\cite{liang2025parameter}. All experiments are conducted on two RTX 4090 GPUs with a total batch size of 4. 

\subsection{Main Results}
In this section, we evaluate PointTPA on scene datasets both with and without a task-specific decoder. The setting with a decoder is primarily evaluated on ScanNet~\cite{dai2017scannet}, while the setting without the decoder better evaluates the feature extraction capability of PEFT methods. For our experiments, we set the prefix~\cite{li2021prefix} token length to 4, the VeRA~\cite{kopiczkovera} rank to 256, the LoRA rank to 32, and the adapter's intermediate dimension to 64. The hyperparameters are shared across all datasets throughout the validation process.

\begin{figure}[!t]
\centering
\includegraphics[width=\linewidth]{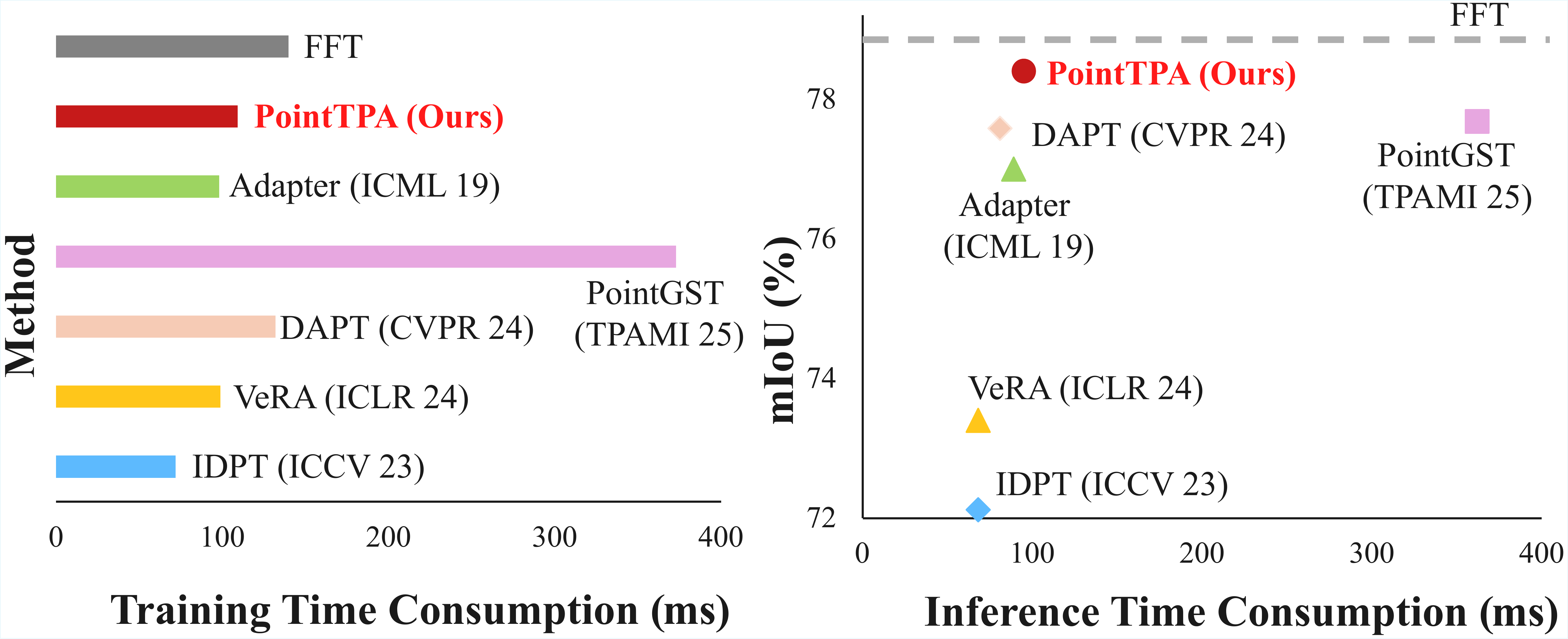}
\caption{Comparison of PEFT methods and FFT on ScanNet~\cite{dai2017scannet} in terms of average training and inference time per sample.}
\label{time}
\vspace{-5pt}
\end{figure}

\textbf{Semantic segmentation without decoder.} 
We evaluate the comprehensive performance and computational efficiency of different approaches, as shown in Tab.~\ref{tab:main_result} and Fig.~\ref{time}. Our PointTPA achieves an optimal balance between accuracy and computational efficiency. While most PEFT methods maintain comparable and low time consumption, they suffer from noticeable performance gaps compared to PointTPA. As for the PointGST, the SOTA method for point cloud analysis, PointTPA operates approximately 4$\times$ faster in both training and inference and reaches 78.4\% mIoU on ScanNet, surpassing PointGST by 0.7\%. On S3DIS, PointTPA further yields improvements of 1.4\% in mAcc and 0.4\% in mIoU over the full fine-tuning reference.

In particular, several PEFT methods fail to achieve strong results. For IDPT, this can be attributed to the fact that IDPT inserts a single prompt token, losing impact on thousands of scene-level tokens. Similarly, VeRA shows marginal gains, as its bottleneck of only two learnable vectors limits its capacity for complex point cloud scenarios. Overall, PEFT methods designed for point clouds outperform general methods, indicating that tailored geometric modules are essential for scene-level understanding.

\begin{table}
  \caption{Results on ScanNet Validation~\cite{dai2017scannet} with decoder.}
  \label{tab:result_with_decoder}
  \footnotesize
  \centering
  \resizebox{\columnwidth}{!}{%
  \begin{tabular}{@{}l*{6}{c}@{}}
    \toprule
    \multirow{2.3}{*}{ } & \multicolumn{2}{c}{\multirow{2.3}{*}{Param. (M)}}& \multicolumn{3}{c}
    {ScanNet Val~\cite{dai2017scannet}}  \\
     \cmidrule(lr){4-6} 
     & &  &mAcc &allAcc&mIoU \\
    \midrule
    \multicolumn{6}{l}{\textit{Full Fine-tuning}}\\
    \midrule
     PTv3-PPT~\cite{wu2024towards}  &\multicolumn{2}{c}{124.8} &86.0 &92.5 &78.6\\
    Sonata (\text{dec.})~\cite{wu2025sonata}  & \multicolumn{2}{c}{124.8}   &86.1 &92.5& 79.4   \\
    \midrule
    \multicolumn{6}{l}{\textit{General PEFT methods}}\\
    \midrule
    Sonata (decoder probing) &\multicolumn{2}{c}{16.3 (13.1\%)}  &85.8 &92.4 &78.6 \\
    + Adapter~\cite{houlsby2019parameter} &\multicolumn{2}{c}{18.2 (14.7\%)} &86.0\dplus{+0.2} &92.4\dtplus{-0.0} &78.8\dplus{+0.2} \\
    + Prefix Tuning~\cite{li2021prefix} &\multicolumn{2}{c}{16.4 (13.5\%)}  &86.0\dplus{+0.2} &92.3\dtplus{-0.1} &78.3\dtplus{-0.3} \\
    + BitFit~\cite{zaken2022bitfit}  &\multicolumn{2}{c}{16.4 (13.5\%)}  &86.5\dplus{+0.7} &92.3\dtplus{-0.1} &78.4\dtplus{-0.2}\\
    + LoRA~\cite{hu2021lora} &\multicolumn{2}{c}{17.2 (13.9\%)}   &85.8\dtplus{-0.0} &92.5\dplus{+0.1} &78.4\dtplus{-0.2} \\
    + VeRA~\cite{kopiczkovera} &\multicolumn{2}{c}{16.4 (13.5\%)} &86.4\dplus{+0.6} &92.4\dtplus{-0.0} &78.5\dtplus{-0.1} \\
    + RandLoRA~\cite{albert2024randlora} &\multicolumn{2}{c}{17.0 (13.7\%)} &86.0\dplus{+0.2} &92.4\dtplus{-0.1}  &78.5\dtplus{-0.1} \\
    \midrule
    \multicolumn{6}{l}{\textit{PEFT methods for point cloud}}\\
    \midrule
    + IDPT~\cite{zha2023instance} &\multicolumn{2}{c}{18.7 (15.1\%)}  &85.8\dtplus{-0.0} &92.4\dtplus{-0.0} &78.4\dtplus{-0.2}\\
    + DAPT~\cite{zhou2024dynamic} &\multicolumn{2}{c}{17.5 (14.1\%)}  &\textbf{86.7}\dplus{+0.9} &92.4\dtplus{-0.0} &79.0\dplus{+0.4}\\
    + PointGST~\cite{liang2025parameter} &\multicolumn{2}{c}{17.4 (14.0\%)} &\textbf{86.7}\dplus{+0.9} &92.4\dtplus{-0.0} &78.8\dplus{+0.2}\\
    + \textbf{PointTPA (\text{ours})} &\multicolumn{2}{c}{17.5 (14.1\%)}  &86.5\dplus{+0.7} &\textbf{92.5}\dplus{+0.1} &\textbf{79.2}\dplus{+0.6}\\
    \bottomrule
  \end{tabular}
  }
  \vspace{-10pt}
\end{table}

\textbf{Segmentation with decoder.} 
To further evaluate the capability of PointTPA, additional experiments are conducted with a task-specific decoder. The decoder consists of four stages, resulting in approximately 16M parameters. As shown in Tab.~\ref{tab:result_with_decoder}, the inclusion of this decoder consistently improves performance for all PEFT methods. General PEFT methods narrow the performance gap with methods for point clouds once the decoder is introduced, because the decoder captures substantial task-specific information. Nevertheless, our PointTPA achieves 92.5\% allAcc and 79.2\% mIoU, nearly matching the FFT reference method, while leading PointGST by 0.4\% in mIoU. These results demonstrate that even with a relatively heavy decoder delivering task-specific information, our inference-time dynamism still provides crucial adaptive signals that further elevate overall scene understanding.

\begin{table*}[ht]
    \centering
    \footnotesize
    \caption{Ablation on the proposed modules, the number of dynamic bases in the parameter base set (Sec.~\ref{subsec:DPP}), and the number of groups in the first encoder stage. ``1*'' denotes no grouping.}
    \label{tab: ablation}
        \begin{subtable}[t]{0.28\linewidth}
            \centering
            \footnotesize
            \setlength{\tabcolsep}{1mm}  
            \captionsetup{margin={22pt, -10pt}}
            \caption{Ablation on SNG and DPP (mIoU).}
            \begin{tabular}{@{}>{\centering\arraybackslash}p{0.9cm}*{4}{>{\centering\arraybackslash}p{1.0cm}}@{}}
                \toprule
                DPP & SNG & ScanNet & \multicolumn{1}{l}{ScanNet++} & S3DIS\\
                \midrule
                \multicolumn{2}{c}{Full Fine-Tuning} & 78.9 & 41.8 & 74.5\\
                \multicolumn{2}{c}{Linear + Adapter} & 76.8 & 39.9 & 73.8\\
                \addlinespace[0.1cm]
                \midrule
                \addlinespace[0.15cm]
                \multicolumn{1}{c@{\hskip -12pt}}{\hspace{-18pt}{\checkmark }}&  & 77.8 & 40.2 & 74.5\\
                \rowcolor{greenbg}
                 \multicolumn{1}{c@{\hskip -12pt}}{\hspace{-18pt}{\checkmark }} &\checkmark &\textbf{78.4} &\multicolumn{1}{c@{\hskip -3pt}}{\hspace{-14pt}{\textbf{40.9}}} & \multicolumn{1}{c@{\hskip 3pt}}{\hspace{3pt}{\textbf{74.9} }}\\
                \bottomrule
            \end{tabular}
            \label{tab:component}
        \end{subtable}%
        \hspace{25pt}
        \begin{subtable}[t]{0.4\linewidth}
            \centering
            \footnotesize
            \setlength{\tabcolsep}{1mm}
            \caption{Ablation on the number of parameter bases.}
             \begin{tabular}{@{}>{\centering\arraybackslash}p{0.8cm}>{\centering\arraybackslash}p{1.5cm}>{\centering\arraybackslash}p{0.6cm}*{3}{>{\centering\arraybackslash}p{0.6cm}}@{}}
                \toprule
                \multicolumn{1}{c}{Number of bases} & Param. (M) & mIoU & mAcc & allAcc\\
                \midrule
                \multicolumn{1}{c}{{[}1, 1, 1, 1, 1{]}} &1.08   &77.7 &85.7 &92.1   \\ 
                \multicolumn{1}{c}{{[}2, 2, 2, 2, 2{]}} &1.16   &78.0 &86.1 &92.1 \\
                \rowcolor{greenbg}
                \multicolumn{1}{c}{{[}4, 4, 2, 2, 2{]}} &1.18   &\textbf{78.4} &\textbf{86.3} &\textbf{92.3}\\
                \multicolumn{1}{c}{{[}4, 4, 4, 4, 4{]}} &1.32   &78.2 &85.8 &92.1 \\
                \multicolumn{1}{c}{{[}8, 8, 4, 4, 4{]}} &1.36   &78.0 &85.8 &92.1  \\
                \bottomrule
            \end{tabular}
            \label{tab:parameter sets}
        \end{subtable}
        \hspace{-8pt}
        \begin{subtable}[t]{0.27\linewidth}
            \centering
            \footnotesize
            \setlength{\tabcolsep}{1mm}
            \caption{Ablation on the number of groups.}
            \begin{tabular}{@{}>{\centering\arraybackslash}p{1cm}*{3}{>{\centering\arraybackslash}p{1cm}}@{}}
                \toprule
                Groups & mIoU & mAcc & allAcc\\
                \midrule
                1* & 77.8 & 85.7 & 92.1\\
                50 & 77.9 & 85.9 & 92.2\\
                100 & 77.9 & 85.6 & 92.1\\
                \rowcolor{greenbg}
                 \multicolumn{1}{c@{\hskip -4pt}}{\hspace{-10pt}{200 }} & \textbf{78.4} & \textbf{86.3} &\multicolumn{1}{c@{\hskip 3pt}}{\hspace{3pt}{\textbf{92.3}}}\\
                400 & 78.0 & 85.9 & 92.0\\
                \bottomrule
            \end{tabular}
            \label{tab:Segmentation}
        \end{subtable}
\end{table*}

\subsection{Ablation Study}
\label{ablation}
This section presents ablation studies evaluating the impact of hyperparameters, architectural designs, and individual components on overall performance. Unless otherwise stated, all experiments are conducted on the ScanNet~\cite{dai2017scannet}. Default settings are masked in \colorbox{greenbg}{green}.

\textbf{Analysis on each component.}
We evaluate the individual contributions of our core modules in Tab.~\ref{tab:component}. Without our modules, the model is linear probing with static adapters, achieving 76.8\% mIoU on ScanNet. Integrating the DPP introduces parameter-efficient dynamism, improving performance by 1.0\% and matching the FFT baseline on S3DIS. Incorporating SNG provides essential spatial guidance for these dynamic weights, yielding an additional 0.6\% gain and outperforming the FFT reference on S3DIS by 0.4\%. Together, these components exhibit a powerful synergy, enabling the base model to dynamically adapt to intricate geometric variations without the heavy computational burden of traditional fine-tuning.

\begin{table*}[t]
    \centering
    \footnotesize
    \caption{Ablation on different settings. We evaluate the effects of (a) scale, (b) DPLayer position, and (c) different space-filling curves in SNG. ``w/o'' denotes the DPP without the DPLayer. Up and Down refers to applying DPLayer to the up and down projectors.}
    \label{other_ablation}
    \begin{subtable}[b]{0.28\linewidth}
        \centering
        \footnotesize
        \setlength{\tabcolsep}{1mm}
        \caption{Ablation on different scaling factors.}
        \begin{tabular}{@{}>{\centering\arraybackslash}p{1cm}*{3}{>{\centering\arraybackslash}p{1cm}}@{}}
            \toprule
            $\bm s$ & mIoU & mAcc & allAcc\\
            \midrule
            0.1 & 78.1 & 86.0 & 92.2\\
            0.5 & 77.8 & 86.2 & 92.1\\
            \rowcolor{greenbg}
            \multicolumn{1}{c@{\hskip -3pt}}{\hspace{-9pt}{1.0}} & \textbf{78.4} & \textbf{86.3} & \multicolumn{1}{c@{\hskip 3pt}}{\hspace{3pt}{\textbf{92.3}}}\\
            2.0 & 77.9 & 86.1 & 92.1\\
            \bottomrule
        \end{tabular}
        \label{tab:scale}
    \end{subtable}
    \hspace{0pt}
    \begin{subtable}[b]{0.38\linewidth}
        \centering
        \footnotesize
        \setlength{\tabcolsep}{1mm}
        \caption{Ablation on different positions of the DPLayer.}
        \begin{tabular}{@{}>{\centering\arraybackslash}p{1cm}>{\centering\arraybackslash}p{1.5cm}>{\centering\arraybackslash}p{1cm}*2{>{\centering\arraybackslash}p{1cm}}@{}}
            \toprule
            \centering{Position} & Params. (M) & mIoU & mAcc & allAcc\\
            \midrule
            w/o & 0.96 & 77.5 & 85.3 & 92.0\\
            Up & 1.07 & 77.4 & 85.4 & 91.9\\
            Both & 1.28 & 77.8 & 85.8 & 92.1\\
            \rowcolor{greenbg}
            \multicolumn{1}{c@{\hskip -3pt}}{\hspace{-8pt}{Down}} & 1.18 & \textbf{78.4} & \textbf{86.3} & \multicolumn{1}{c@{\hskip 3pt}}{\hspace{3pt}{\textbf{92.3}}}\\
            \bottomrule
        \end{tabular}
        \label{tab:dpl}
    \end{subtable}
    \hspace{0pt}
    \begin{subtable}[b]{0.28\linewidth}
    \centering
    \footnotesize
        \setlength{\tabcolsep}{1mm}
        \caption{Comparison of space-filling curves.}
        \begin{tabular}{@{}>{\centering\arraybackslash}p{1cm}>{\centering\arraybackslash}p{1.5cm}>{\centering\arraybackslash}p{1.5cm}@{}}
        \toprule
        Method & mIoU & mAcc \\
        \midrule
        w/o SFC & 77.8 & 85.7\\
        Hilbert & 78.1 & 85.8 \\
        Z-order & 78.2 & 86.1 \\
        \rowcolor{greenbg} 
        \multicolumn{1}{c@{\hskip -4pt}}{\hspace{-10pt}{Mixed}} & \textbf{78.4} & \multicolumn{1}{c@{\hskip 4pt}}{\hspace{4pt}{\textbf{86.3}}} \\
        \bottomrule
    \end{tabular}
    \label{tab:curve_compare}
    \end{subtable}
\end{table*}

\textbf{Analysis on parameter base set.}
The DPP module provides the necessary bases for generating dynamic weights. We evaluate configurations with varying numbers of bases, denoted as [$n_1, n_2, \dots$], where $n_i$ represents the number of bases in the $i$-th stage. Because point resolution decreases with the network stage while the attention blocks use fixed input sizes within each stage, we keep the base count constant within a stage and vary it across stages to balance resolution and computational cost. As shown in Tab.~\ref{tab:parameter sets}, initially increasing the bases boosts performance by introducing greater dynamism for structural adaptation. Consequently, the configuration [4, 4, 2, 2, 2] achieves the optimal trade-off between parameter efficiency and performance. Conversely, heavier setups lead to performance degradation. The optimal setting outperforms the heavier [8, 8, 4, 4, 4] setup by 0.4\%. We attribute this degradation to the overabundance of bases, which complicates convergence. This confirms that while too few bases restrict the diversity of linear combinations, an overabundance induces training instability, making our stage-aware configuration the most effective design.

\textbf{Analysis on the number of groups.}
The number of groups is a core hyperparameter in SNG, directly determining the precision of the local geometric information.  We conduct experiments across various settings. Specifically, the number of groups is progressively halved across stages. Because token sequences shorten in deeper stages, a fixed number of groups would induce excessive sparsity. Tab.~\ref{tab:Segmentation} demonstrates that the number of groups significantly affects performance. In general, increasing the number of groups brings better results as a larger group number forces each group to contain fewer points and focus on more fine-grained geometric structures. However, when the number reaches 400, it results in a significant performance drop with a 0.4\% decrease in mIoU and mAcc compared to 200. We attribute this decline to an insufficient number of tokens per group, resulting in a lack of geometric information.

\textbf{Analysis on the scaling factor.}
The scaling factor $s$ in the DPP branch is crucial for effectively merging backbone features with dynamic representations. As shown in Tab.~\ref{tab:scale}, setting $s=1.0$  improves both mIoU and mAcc by 0.3\% over the conventional $s=0.1$, and consistently achieves better results than other evaluated scales. Therefore, we use $s=1.0$ in all subsequent experiments.

\textbf{Analysis on the position of DPLayer.}
Each DPP consists of a down-projection layer and an up-projection layer to sequentially compress and expand features. We conduct ablation experiments to analyze the effect of applying the DPLayer at different positions. Tab.~\ref{tab:dpl} shows that applying the DPLayer solely to the down-projection step maximizes model performance, achieving a 0.9\% mIoU improvement over the baseline. While replacing both layers remains superior to the baseline, it incurs a 0.6\% mIoU drop compared to our default down-projection configuration. These results indicate that dynamic capacity is most effectively utilized during initial feature extraction, whereas applying it universally introduces unnecessary complexities.

\textbf{Analysis on space-filling curves (SFC).} SFC fundamentally transforms irregular 3D point clouds into predictable, structured sequences. This transformation preserves crucial spatial locality, allowing the model to group points into meaningful geometric patches rather than random subsets. To quantify this impact, we evaluate several curve variants. Tab.~\ref{tab:curve_compare} demonstrates that all curve variants consistently improve performance over the unordered method. Therefore, we naturally adopt the mixed strategy from the PTv3 to maximize this geometric advantage.

\textbf{Analysis on the inserting position of DPP.} Although dynamic parameters increase flexibility, an excessive amount can hinder model optimization. To find the optimal balance between representational capacity and training stability, we evaluate various DPP insertion strategies (Tab.~\ref{tab:insert}). Notably, a dense DPP configuration introduces redundant parameters and degrades performance, reducing mIoU by 0.6\% and allAcc by 0.2\%. This can be attributed to the instability caused by too many dynamic network parameters, which makes convergence difficult. Other settings above also fall short of the default setup.

\begin{table}
  \footnotesize
  \caption{Ablation on the insertion strategy of our DPP. ``Dense'' denotes applying DPP in every block.}
  \label{tab:insert}
  \centering
  \begin{tabular}{@{}p{1.2cm}*{3}{>{\centering\arraybackslash}p{1.1cm}}@{}}
    \toprule
    \multicolumn{1}{l}{Inserting Strategy}   & mIoU & mAcc & allAcc\\
    \midrule
    \multicolumn{1}{l}{Dense} &77.8 &\textbf{86.6} & 92.1\\
    \multicolumn{1}{l}{Every two blocks} &77.9 &85.8 &92.1\\
    \multicolumn{1}{l}{Every three blocks} &77.7 &85.7 & 92.1\\
    \multicolumn{1}{l}{Last block}  &77.6 &85.4 &91.2 \\
    \multicolumn{1}{l}{First block in every stage} &78.0 &85.8 &92.1\\
    \rowcolor{greenbg}
    \multicolumn{1}{l}{Last block in every stage} &\textbf{78.4} &86.3 & \multicolumn{1}{c@{\hskip 6pt}}{\hspace{6pt}{\textbf{92.3}}}\\ 
    \bottomrule
  \end{tabular}
\end{table}

\begin{figure}[t]
\centering
\includegraphics[width=\linewidth]{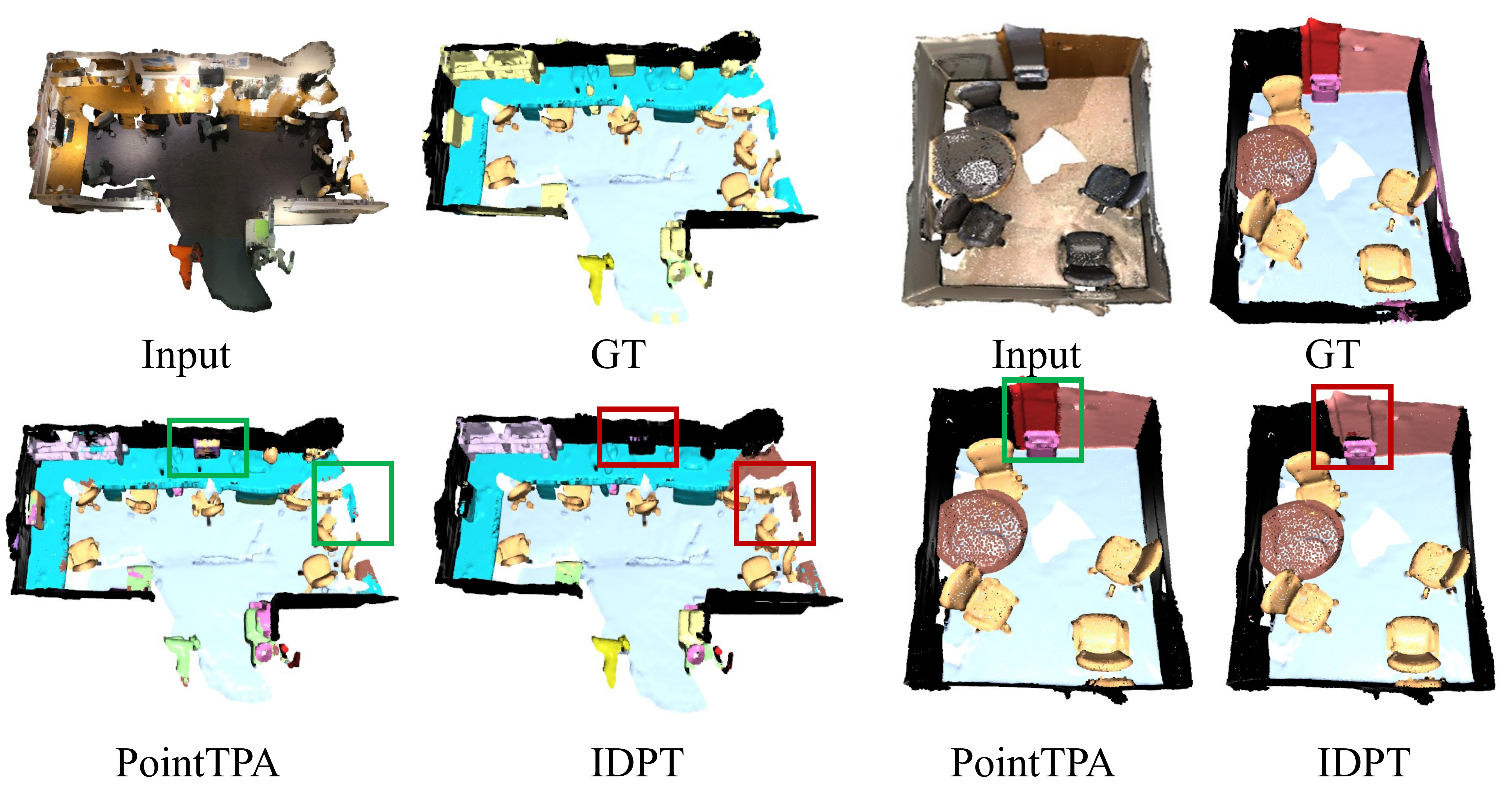}
\caption{Qualitative comparison against IDPT~\cite{zha2023instance}. Green and red boxes indicate correct and incorrect segmentations, respectively, with GT denoting the ground truth.}
\label{fig:vis}
\vspace{-8pt}
\end{figure}

\subsection{Qualitative Analysis}
We conduct qualitative experiments on the ScanNet~\cite{dai2017scannet} dataset. As Fig.~\ref{fig:vis} shows, PointTPA produces high-quality segmentation on common objects, highlighting its steady fundamental performance. Furthermore, PointTPA accurately segments wall-adjacent objects, such as doors and televisions, in complex indoor scenes. In contrast, IDPT exhibits severe confusion and often misses these instances. This indicates that PointTPA effectively discerns subtle geometric variations, allowing for highly accurate fine-grained structure segmentation.

\begin{figure}[t]
\centering
\includegraphics[width=\linewidth]{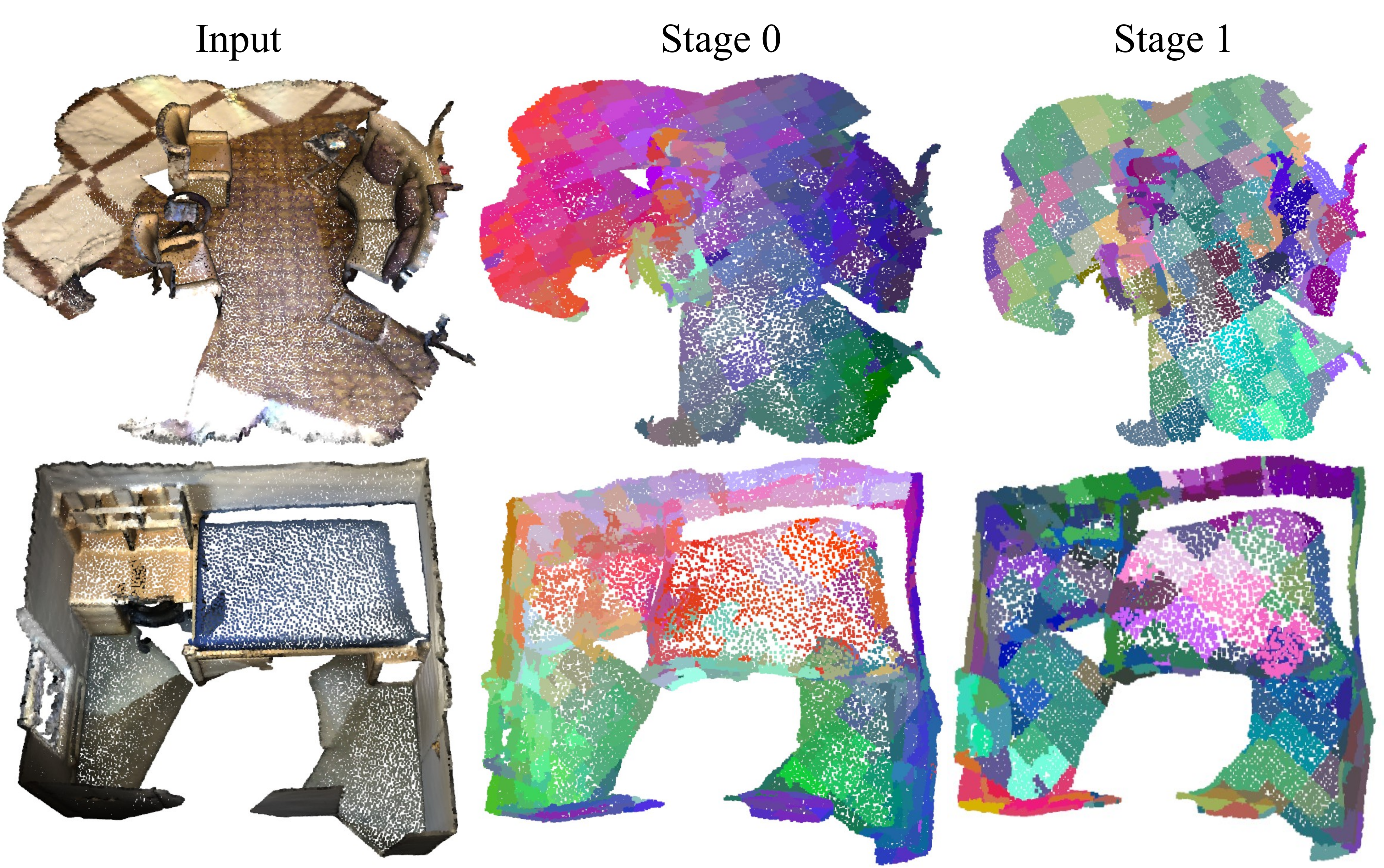}
\caption{Visualization of the similarity of dynamic weights.}
\label{dy}
\vspace{-4pt}
\end{figure}

To further evaluate the dynamic learning ability of PointTPA, we visualize the similarity of dynamically generated projection weights across different groups and stages, with similar colors indicating higher similarity. As shown in Fig.~\ref{dy}, PointTPA produces distinct color patterns across different point groups, indicating significant variance in the generated dynamic weights. This shows that PointTPA successfully generates distinct dynamic weights for different patches, enabling it to extract richer structural features for better scene understanding.
% \vspace{3pt}

\section{Conclusion}
In this study, we propose PointTPA, a dynamic PEFT method for 3D scene understanding. During training, the pretrained backbone remains frozen while the SNG groups tokens based on local geometries. Guided by these geometric groups, the DPP adaptively refines the representation of distinct local structures. During inference, the DPP produces adaptive projection weights for local token groups by generating specific routing coefficients. This design enables PointTPA to generate high-quality local representations for the backbone. While our method demonstrates notable efficiency and competitiveness, there is still room for improvement. Nevertheless, our dynamic inference approach presents an interesting direction for further study.

{
    \small
    \bibliographystyle{ieeenat_fullname}
    \bibliography{main}

@string{CVPR = "Proc. IEEE Conf. Comput. Vis. Pattern Recognit."}

@string{ICCV = "Proc. IEEE Int. Conf. Comput. Vis."}

@string{ECCV = "Proc. Eur. Conf. Comput. Vis."}

@string{NeurIPS = "Proc. Adv. Neural Inf. Process. Syst."}

@string{ICML = "Proc. Int. Conf. Mach. Learn."}

@string{IJCAI = "Proc. Int. Joint Conf. Artif. Intell."}

@string{AAAI = "Proc. AAAI Conf. Artif. Intell."}

@string{ICLR = "Proc. Int. Conf. Learn. Representations"}

@string{ICASSP = "Proc. Int. Conf. Acoustics, Speech, Signal Process."}

@string{ICRA   = "Proc. IEEE Int. Conf. Robotics Automation"}

@string{ACL = "Proc. Annual Meeting of the Association for Computational Linguistics"}

@string{TPAMI = "IEEE Trans. Pattern Anal. Mach. Intell."}

@string{TOG = "ACM Trans. ON Graphics"}

@string{TIV = "IEEE Trans. Intell. Vehicles"}

@article{guo2020deep,
  title={Deep learning for 3d point clouds: A survey},
  author={Guo, Yulan and Wang, Hanyun and Hu, Qingyong and Liu, Hao and Liu, Li and Bennamoun, Mohammed},
  journal=TPAMI,
  volume={43},
  number={12},
  pages={4338--4364},
  year={2020},
}

@article{lu2022transformers,
  title={Transformers in 3d point clouds: A survey},
  author={Lu, Dening and Xie, Qian and Wei, Mingqiang and Gao, Kyle and Xu, Linlin and Li, Jonathan},
  journal={arXiv preprint arXiv:2205.07417},
  year={2022}
}

@inproceedings{afham2022crosspoint,
  title={Crosspoint: Self-supervised cross-modal contrastive learning for 3d point cloud understanding},
  author={Afham, Mohamed and Dissanayake, Isuru and Dissanayake, Dinithi and Dharmasiri, Amaya and Thilakarathna, Kanchana and Rodrigo, Ranga},
  booktitle=CVPR,
  pages={9902--9912},
  year={2022}
}

@inproceedings{chen2023clip2scene,
  title={Clip2scene: Towards label-efficient 3d scene understanding by clip},
  author={Chen, Runnan and Liu, Youquan and Kong, Lingdong and Zhu, Xinge and Ma, Yuexin and Li, Yikang and Hou, Yuenan and Qiao, Yu and Wang, Wenping},
  booktitle=CVPR,
  pages={7020--7030},
  year={2023}
}

@inproceedings{zhang2022point,
  title={Point-m2ae: multi-scale masked autoencoders for hierarchical point cloud pre-training},
  author={Zhang, Renrui and Guo, Ziyu and Gao, Peng and Fang, Rongyao and Zhao, Bin and Wang, Dong and Qiao, Yu and Li, Hongsheng},
  booktitle=NeurIPS,
  volume={35},
  pages={27061--27074},
  year={2022}
}

@inproceedings{choy20194d,
  title={4d spatio-temporal convnets: Minkowski convolutional neural networks},
  author={Choy, Christopher and Gwak, JunYoung and Savarese, Silvio},
  booktitle=CVPR,
  pages={3075--3084},
  year={2019}
}

@inproceedings{liang2024pointmamba,
  title={PointMamba: A Simple State Space Model for Point Cloud Analysis},
  author={Liang, Dingkang and Zhou, Xin and Xu, Wei and Zhu, Xingkui and Zou, Zhikang and Ye, Xiaoqing and Tan, Xiao and Bai, Xiang},
  booktitle=NeurIPS,
  volume = {37},
  year={2024},
  pages = {32653--32677}
}

@inproceedings{qi2017pointnet,
  title={Pointnet: Deep learning on point sets for 3d classification and segmentation},
  author={Qi, Charles R and Su, Hao and Mo, Kaichun and Guibas, Leonidas J},
  booktitle=CVPR,
  pages={652--660},
  year={2017}
}

@inproceedings{qi2017pointnet++,
  title={Pointnet++: Deep hierarchical feature learning on point sets in a metric space},
  author={Qi, Charles Ruizhongtai and Yi, Li and Su, Hao and Guibas, Leonidas J},
  booktitle=NeurIPS,
  volume={30},
  year={2017},
  pages={5105--5114},
}

@inproceedings{zhou2018voxelnet,
  title={Voxelnet: End-to-end learning for point cloud based 3d object detection},
  author={Zhou, Yin and Tuzel, Oncel},
  booktitle=CVPR,
  pages={4490--4499},
  year={2018}
}

@inproceedings{mao2021voxel,
  title={Voxel transformer for 3d object detection},
  author={Mao, Jiageng and Xue, Yujing and Niu, Minzhe and Bai, Haoyue and Feng, Jiashi and Liang, Xiaodan and Xu, Hang and Xu, Chunjing},
  booktitle=ICCV,
  pages={3164--3173},
  year={2021}
}

@inproceedings{su2015multi,
  title={Multi-view convolutional neural networks for 3d shape recognition},
  author={Su, Hang and Maji, Subhransu and Kalogerakis, Evangelos and Learned-Miller, Erik},
  booktitle=ICCV,
  pages={945--953},
  year={2015}
}

@inproceedings{xie2020pointcontrast,
  title={Pointcontrast: Unsupervised pre-training for 3d point cloud understanding},
  author={Xie, Saining and Gu, Jiatao and Guo, Demi and Qi, Charles R and Guibas, Leonidas and Litany, Or},
  booktitle=ECCV,
  pages={574--591},
  year={2020},
}

@inproceedings{wu2023masked,
  title={Masked scene contrast: A scalable framework for unsupervised 3d representation learning},
  author={Wu, Xiaoyang and Wen, Xin and Liu, Xihui and Zhao, Hengshuang},
  booktitle=CVPR,
  pages={9415--9424},
  year={2023}
}

@inproceedings{zhao2021point,
  title={Point transformer},
  author={Zhao, Hengshuang and Jiang, Li and Jia, Jiaya and Torr, Philip HS and Koltun, Vladlen},
  booktitle=ICCV,
  pages={16259--16268},
  year={2021}
}

@article{guo2021pct,
  title={Pct: Point cloud transformer},
  author={Guo, Meng-Hao and Cai, Jun-Xiong and Liu, Zheng-Ning and Mu, Tai-Jiang and Martin, Ralph R and Hu, Shi-Min},
  journal={Computational visual media},
  volume={7},
  number={2},
  pages={187--199},
  year={2021},
}

@article{wang2019dynamic,
  title={Dynamic graph cnn for learning on point clouds},
  author={Wang, Yue and Sun, Yongbin and Liu, Ziwei and Sarma, Sanjay E and Bronstein, Michael M and Solomon, Justin M},
  journal=TOG,
  volume={38},
  number={5},
  pages={1--12},
  year={2019},
}

@article{zha2024pre,
  title={Pre-training Point Cloud Compact Model with Partial-aware Reconstruction},
  author={Zha, Yaohua and Wang, Yanzi and Dai, Tao and Xia, Shu-Tao},
  journal={arXiv preprint arXiv:2407.09344},
  year={2024}
}

@inproceedings{zha2025pma,
  title={Pma: Towards parameter-efficient point cloud understanding via point mamba adapter},
  author={Zha, Yaohua and Wang, Yanzi and Guo, Hang and Wang, Jinpeng and Dai, Tao and Chen, Bin and Ouyang, Zhihao and Yuerong, Xue and Chen, Ke and Xia, Shu-Tao},
  booktitle=CVPR,
  pages={16976--16986},
  year={2025}
}

@inproceedings{zha2023sfr,
  title={Sfr: Semantic-aware feature rendering of point cloud},
  author={Zha, Yaohua and Li, Rongsheng and Dai, Tao and Xiong, Jianyu and Wang, Xin and Xia, Shu-Tao},
  booktitle=ICASSP,
  pages={1--5},
  year={2023},
}

@inproceedings{zha2024lcm,
  title={Lcm: Locally constrained compact point cloud model for masked point modeling},
  author={Zha, Yaohua and Li, Naiqi and Wang, Yanzi and Dai, Tao and Guo, Hang and Chen, Bin and Wang, Zhi and Ouyang, Zhihao and Xia, Shu-Tao},
  journal=NeurIPS,
  volume={37},
  pages={104816--104842},
  year={2024},
}

@inproceedings{zha2024towards,
  title={Towards compact 3d representations via point feature enhancement masked autoencoders},
  author={Zha, Yaohua and Ji, Huizhen and Li, Jinmin and Li, Rongsheng and Dai, Tao and Chen, Bin and Wang, Zhi and Xia, Shu-Tao},
  booktitle=AAAI,
  volume={38},
  number={7},
  pages={6962--6970},
  year={2024}
}

@inproceedings{wu2022point,
  title={Point transformer v2: Grouped vector attention and partition-based pooling},
  author={Wu, Xiaoyang and Lao, Yixing and Jiang, Li and Liu, Xihui and Zhao, Hengshuang},
  booktitle=NeurIPS,
  volume={35},
  pages={33330--33342},
  year={2022}
}

@inproceedings{wu2024point,
  title={Point transformer v3: Simpler faster stronger},
  author={Wu, Xiaoyang and Jiang, Li and Wang, Peng-Shuai and Liu, Zhijian and Liu, Xihui and Qiao, Yu and Ouyang, Wanli and He, Tong and Zhao, Hengshuang},
  booktitle=CVPR,
  pages={4840--4851},
  year={2024}
}

@inproceedings{zhou2024dynamic,
  title={Dynamic adapter meets prompt tuning: Parameter-efficient transfer learning for point cloud analysis},
  author={Zhou, Xin and Liang, Dingkang and Xu, Wei and Zhu, Xingkui and Xu, Yihan and Zou, Zhikang and Bai, Xiang},
  booktitle=CVPR,
  pages={14707--14717},
  year={2024}
}

@inproceedings{pang2022masked,
  title={Masked autoencoders for point cloud self-supervised learning},
  author={Pang, Yatian and Wang, Wenxiao and Tay, Francis EH and Liu, Wei and Tian, Yonghong and Yuan, Li},
  booktitle=ECCV,
  pages={604--621},
  year={2022},
}

@inproceedings{zha2023instance,
  title={Instance-aware dynamic prompt tuning for pre-trained point cloud models},
  author={Zha, Yaohua and Wang, Jinpeng and Dai, Tao and Chen, Bin and Wang, Zhi and Xia, Shu-Tao},
  booktitle=ICCV,
  pages={14161--14170},
  year={2023}
}

@inproceedings{sun2024parameter,
  title={Parameter-efficient prompt learning for 3d point cloud understanding},
  author={Sun, Hongyu and Wang, Yongcai and Chen, Wang and Deng, Haoran and Li, Deying},
  booktitle=ICRA,
  pages={9478--9486},
  year={2024},
}

@article{liang2025parameter,
  title={Parameter-efficient fine-tuning in spectral domain for point cloud learning},
  author={Liang, Dingkang and Feng, Tianrui and Zhou, Xin and Zhang, Yumeng and Zou, Zhikang and Bai, Xiang},
  journal=TPAMI,
  year={2025},
  volume={47},
  number={12},
  pages={10949-10966}
}

@inproceedings{yu2022point,
  title={Point-bert: Pre-training 3d point cloud transformers with masked point modeling},
  author={Yu, Xumin and Tang, Lulu and Rao, Yongming and Huang, Tiejun and Zhou, Jie and Lu, Jiwen},
  booktitle=CVPR,
  pages={19313--19322},
  year={2022}
}

@inproceedings{wu2025sonata,
  title={Sonata: Self-supervised learning of reliable point representations},
  author={Wu, Xiaoyang and DeTone, Daniel and Frost, Duncan and Shen, Tianwei and Xie, Chris and Yang, Nan and Engel, Jakob and Newcombe, Richard and Zhao, Hengshuang and Straub, Julian},
  booktitle=CVPR,
  pages={22193--22204},
  year={2025}
}

@inproceedings{chen2023voxelnext,
  title={Voxelnext: Fully sparse voxelnet for 3d object detection and tracking},
  author={Chen, Yukang and Liu, Jianhui and Zhang, Xiangyu and Qi, Xiaojuan and Jia, Jiaya},
  booktitle=CVPR,
  pages={21674--21683},
  year={2023}
}

@inproceedings{tang2024point,
  title={Point-peft: Parameter-efficient fine-tuning for 3d pre-trained models},
  author={Tang, Yiwen and Zhang, Ray and Guo, Zoey and Ma, Xianzheng and Zhao, Bin and Wang, Zhigang and Wang, Dong and Li, Xuelong},
  booktitle=AAAI,
  volume={38},
  number={6},
  pages={5171--5179},
  year={2024}
}

@inproceedings{albert2024randlora,
  title={RandLoRA: full rank parameter-efficient fine-tuning of large models},
  author={Albert, Paul and Zhang, Frederic Z and Rodriguez-Opazo, Cristian and Saratchandran, Hemanth and Hengel, Anton van den and Abbasnejad, Ehsan},
  journal=ICLR,
  year={2024}
}

@inproceedings{hu2021lora,
  title={LoRA: Low-Rank Adaptation of Large Language Models},
  author={Hu, Edward J and Wallis, Phillip and Allen-Zhu, Zeyuan and Li, Yuanzhi and Wang, Shean and Wang, Lu and Chen, Weizhu and others},
  booktitle=ICLR,
  year={2022}
}

@inproceedings{houlsby2019parameter,
  title={Parameter-efficient transfer learning for NLP},
  author={Houlsby, Neil and Giurgiu, Andrei and Jastrzebski, Stanislaw and Morrone, Bruna and De Laroussilhe, Quentin and Gesmundo, Andrea and Attariyan, Mona and Gelly, Sylvain},
  booktitle=ICML,
  pages={2790--2799},
  year={2019},
}

@inproceedings{jia2022visual,
  title={Visual prompt tuning},
  author={Jia, Menglin and Tang, Luming and Chen, Bor-Chun and Cardie, Claire and Belongie, Serge and Hariharan, Bharath and Lim, Ser-Nam},
  booktitle=ECCV,
  pages={709--727},
  year={2022},
}

@inproceedings{valipour2023dylora,
  title={DyLoRA: Parameter-efficient tuning of pre-trained models using dynamic search-free low-rank adaptation},
  author={Valipour, Mojtaba and Rezagholizadeh, Mehdi and Kobyzev, Ivan and Ghodsi, Ali},
  booktitle={Proceedings of the 17th Conference of the European Chapter of the Association for Computational Linguistics},
  pages={3274--3287},
  year={2023}
}

@article{fei2025parameter,
  title={Parameter Efficient Point Cloud Prompt Tuning for Unified Point Cloud Understanding},
  author={Fei, Ben and Liu, Liwen and Yang, Weidong and Li, Zhijun and Chen, Wen-Ming and Ma, Lipeng},
  journal=TIV,
  volume={10},
  number={1},
  pages={255--271},
  year={2025}
}

@inproceedings{tang2024any2point,
  title={Any2point: Empowering any-modality large models for efficient 3d understanding},
  author={Tang, Yiwen and Zhang, Ray and Liu, Jiaming and Guo, Zoey and Zhao, Bin and Wang, Zhigang and Gao, Peng and Li, Hongsheng and Wang, Dong and Li, Xuelong},
  booktitle=ECCV,
  pages={456--473},
  year={2024},
}

@inproceedings{li2021prefix,
  title={Prefix-Tuning: Optimizing Continuous Prompts for Generation},
  author={Li, Xiang Lisa and Liang, Percy},
  booktitle=ACL,
  pages={4582--4597},
  year={2021}
}

@inproceedings{zaken2022bitfit,
  title={Bitfit: Simple parameter-efficient fine-tuning for transformer-based masked language-models},
  author={Zaken, Elad Ben and Goldberg, Yoav and Ravfogel, Shauli},
  booktitle=ACL,
  pages={1--9},
  year={2022}
}

@misc{pointcept2023,
    title={Pointcept: A Codebase for Point Cloud Perception Research},
    author={Pointcept Contributors},
    howpublished = {\url{https://github.com/Pointcept/Pointcept}},
    year={2023}
}

@inproceedings{armeni20163d,
  title={3d semantic parsing of large-scale indoor spaces},
  author={Armeni, Iro and Sener, Ozan and Zamir, Amir R and Jiang, Helen and Brilakis, Ioannis and Fischer, Martin and Savarese, Silvio},
  booktitle=CVPR,
  pages={1534--1543},
  year={2016}
}

@inproceedings{dai2017scannet,
  title={Scannet: Richly-annotated 3d reconstructions of indoor scenes},
  author={Dai, Angela and Chang, Angel X and Savva, Manolis and Halber, Maciej and Funkhouser, Thomas and Nie{\ss}ner, Matthias},
  booktitle=CVPR,
  pages={5828--5839},
  year={2017}
}

@inproceedings{rozenberszki2022language,
  title={Language-grounded indoor 3d semantic segmentation in the wild},
  author={Rozenberszki, David and Litany, Or and Dai, Angela},
  booktitle=ECCV,
  pages={125--141},
  year={2022},
}

@inproceedings{yeshwanth2023scannet++,
  title={Scannet++: A high-fidelity dataset of 3d indoor scenes},
  author={Yeshwanth, Chandan and Liu, Yueh-Cheng and Nie{\ss}ner, Matthias and Dai, Angela},
  booktitle=ICCV,
  pages={12--22},
  year={2023}
}

@inproceedings{kopiczkovera,
  title={VeRA: Vector-based Random Matrix Adaptation},
  author={Kopiczko, Dawid Jan and Blankevoort, Tijmen and Asano, Yuki M},
  booktitle=ICLR,
  year={2024}
}

@inproceedings{jiang2020pointgroup,
  title={Pointgroup: Dual-set point grouping for 3d instance segmentation},
  author={Jiang, Li and Zhao, Hengshuang and Shi, Shaoshuai and Liu, Shu and Fu, Chi-Wing and Jia, Jiaya},
  booktitle=CVPR,
  pages={4867--4876},
  year={2020}
}

@inproceedings{zhang2022pointclip,
  title={Pointclip: Point cloud understanding by clip},
  author={Zhang, Renrui and Guo, Ziyu and Zhang, Wei and Li, Kunchang and Miao, Xupeng and Cui, Bin and Qiao, Yu and Gao, Peng and Li, Hongsheng},
  booktitle=CVPR,
  pages={8552--8562},
  year={2022}
}

@inproceedings{qian2022pointnext,
    title={Pointnext: Revisiting pointnet++ with improved training and scaling strategies},
    author={Qian, Guocheng and Li, Yuchen and Peng, Houwen and Mai, Jinjie and Hammoud, Hasan and Elhoseiny, Mohamed and Ghanem, Bernard},
    booktitle =NeurIPS,  
    volume = {35},
    pages = {23192--23204},
    year = {2022}
}

@inproceedings{zhu2023pointclip,
  title={Pointclip v2: Prompting clip and gpt for powerful 3d open-world learning},
  author={Zhu, Xiangyang and Zhang, Renrui and He, Bowei and Guo, Ziyu and Zeng, Ziyao and Qin, Zipeng and Zhang, Shanghang and Gao, Peng},
  booktitle=ICCV,
  pages={2639--2650},
  year={2023}
}

@inproceedings{park2023self,
  title={Self-positioning point-based transformer for point cloud understanding},
  author={Park, Jinyoung and Lee, Sanghyeok and Kim, Sihyeon and Xiong, Yunyang and Kim, Hyunwoo J},
  booktitle=CVPR,
  pages={21814--21823},
  year={2023}
}

@inproceedings{liu2020closer,
  title={A closer look at local aggregation operators in point cloud analysis},
  author={Liu, Ze and Hu, Han and Cao, Yue and Zhang, Zheng and Tong, Xin},
  booktitle=ECCV,
  pages={326--342},
  year={2020},
}

@inproceedings{xiang2021walk,
  title={Walk in the cloud: Learning curves for point clouds shape analysis},
  author={Xiang, Tiange and Zhang, Chaoyi and Song, Yang and Yu, Jianhui and Cai, Weidong},
  booktitle=ICCV,
  pages={915--924},
  year={2021}
}

@inproceedings{Zhang_2023_CVPR,
    title={Starting From Non-Parametric Networks for 3D Point Cloud Analysis},
    author={Zhang, Renrui and Wang, Liuhui and Wang, Yali and Gao, Peng and Li, Hongsheng and Shi, Jianbo},
    booktitle=CVPR,
    pages={5344--5353},
    year={2023}
}

@inproceedings{hou2021exploring,
  title={Exploring data-efficient 3d scene understanding with contrastive scene contexts},
  author={Hou, Ji and Graham, Benjamin and Nie{\ss}ner, Matthias and Xie, Saining},
  booktitle=CVPR,
  pages={15587--15597},
  year={2021}
}

@inproceedings{wu2024towards,
  title={Towards large-scale 3d representation learning with multi-dataset point prompt training},
  author={Wu, Xiaoyang and Tian, Zhuotao and Wen, Xin and Peng, Bohao and Liu, Xihui and Yu, Kaicheng and Zhao, Hengshuang},
  booktitle=CVPR,
  pages={19551--19562},
  year={2024}
}

@article{fan2023super,
  title={Super sparse 3d object detection},
  author={Fan, Lue and Yang, Yuxue and Wang, Feng and Wang, Naiyan and Zhang, Zhaoxiang},
  journal=TPAMI,
  volume={45},
  number={10},
  pages={12490--12505},
  year={2023},
}

@inproceedings{li2023dds3d,
  title={DDS3D: Dense Pseudo-Labels with Dynamic Threshold for Semi-Supervised 3D Object Detection},
  author={Li, Jingyu and Liu, Zhe and Hou, Jinghua and Liang, Dingkang},
  booktitle=ICRA,
  pages={9245--9252},
  year={2023},
}

@inproceedings{li2026imagidrive,
    author ={Li, Jingyu and Zhang, Bozhou and Jin, Xin and Deng, Jiankang and Zhu, Xiatian and Zhang, Li},
    title ={ImagiDrive: A Unified Imagination-and-Planning Framework for Autonomous Driving},
    booktitle=ICRA,
    year={2026}
}

@inproceedings{li2026geoteacher,
  title={GeoTeacher: Geometry-Guided Semi-Supervised 3D Object Detection},
  author={Li, Jingyu and Zhao, Xiaolong and Liu, Zhe and Wu, Wenxiao and Zhang, Li},
  booktitle =ICRA,
  year={2026}
}

@inproceedings{li2026sgdrive,
    author = {Li, Jingyu and Wu, Junjie and Hu, Dongnan and Huang, Xiangkai and Sun, Bin and Hao, Zhihui and Lang, Xianpeng and Zhu, Xiatian and Zhang, Li},
    title ={SGDrive: Scene-to-Goal Hierarchical World Cognition for Autonomous Driving},
    booktitle =CVPR,
    year={2026}
}

@inproceedings{xu2024unified,
  title={A unified framework for 3D scene understanding},
  author={Xu, Wei and Shi, Chunsheng and Tu, Sifan and Zhou, Xin and Liang, Dingkang and Bai, Xiang},
  booktitle=NeurIPS,
  volume = {37},
  pages = {59468--59490},
  year={2024}
}

@inproceedings{zhang2023simple,
  title={A simple vision transformer for weakly semi-supervised 3d object detection},
  author={Zhang, Dingyuan and Liang, Dingkang and Zou, Zhikang and Li, Jingyu and Ye, Xiaoqing and Liu, Zhe and Tan, Xiao and Bai, Xiang},
  booktitle=ICCV,
  pages={8373-8383},
  year={2023}
}

@inproceedings{zhou2025hermes,
  title={Hermes: A unified self-driving world model for simultaneous 3d scene understanding and generation},
  author={Zhou, Xin and Liang, Dingkang and Tu, Sifan and Chen, Xiwu and Ding, Yikang and Zhang, Dingyuan and Tan, Feiyang and Zhao, Hengshuang and Bai, Xiang},
  booktitle=ICCV,
  pages={27817--27827},
  year={2025}
}

@inproceedings{liang2025seeing,
    author = {Liang, Dingkang and Zhang, Dingyuan and Zhou, Xin and Tu, Sifan and Feng, Tianrui and Li, Xiaofan and Zhang, Yumeng and Du, Mingyang and Tan, Xiao and Bai, Xiang},
    title ={UniFuture: A 4D Driving World Model for Future Generation and Perception} ,
    booktitle =ICRA,
    year={2026}
}

@inproceedings{zha2025point,
  title={Point cloud mixture-of-domain-experts model for 3D self-supervised learning},
  author={Zha, Yaohua and Dai, Tao and Guo, Hang and Wang, Yanzi and Chen, Bin and Chen, Ke and Xia, Shu-Tao},
  booktitle=IJCAI,
  pages={2332--2340},
  year={2025}
}

@inproceedings{liang2026cook,
  title={Cook and clean together: Teaching embodied agents for parallel task execution},
  author={Liang, Dingkang and Zhang, Cheng and Xu, Xiaopeng and Ju, Jianzhong and Luo, Zhenbo and Bai, Xiang},
  booktitle=AAAI,
  volume={40},
  number={22},
  pages={18415--18424},
  year={2026}
}

@inproceedings{fu2025orion,
  title={Orion: A holistic end-to-end autonomous driving framework by vision-language instructed action generation},
  author={Fu, Haoyu and Zhang, Diankun and Zhao, Zongchuang and Cui, Jianfeng and Liang, Dingkang and Zhang, Chong and Zhang, Dingyuan and Xie, Hongwei and Wang, Bing and Bai, Xiang},
  booktitle=ICCV,
  pages={24823--24834},
  year={2025}
}

@article{fu2025minddrive,
  title={MindDrive: A Vision-Language-Action Model for Autonomous Driving via Online Reinforcement Learning},
  author={Fu, Haoyu and Zhang, Diankun and Zhao, Zongchuang and Cui, Jianfeng and Xie, Hongwei and Wang, Bing and Chen, Guang and Liang, Dingkang and Bai, Xiang},
  journal={arXiv preprint arXiv:2512.13636},
  year={2025}
}

@article{liang2025sood++,
  title={Sood++: Leveraging unlabeled data to boost oriented object detection},
  author={Liang, Dingkang and Hua, Wei and Shi, Chunsheng and Zou, Zhikang and Ye, Xiaoqing and Bai, Xiang},
  journal=TPAMI,
  year={2025},
  pages={840-858},
  volume={48},
  number={1},
}
}

\setcounter{figure}{0}
\setcounter{table}{0}
\setcounter{section}{0}
\renewcommand{\thesection}{S\arabic{section}}

\clearpage
\supptitle{Supplementary Material}
\section{Additional Experiments}
\label{sec:additional experiments}

\subsection{Analysis on Different Rank}
One of our core hyperparameters is the rank $r$, which controls the feature dimension of the projection and has a significant impact on both the number of parameters and feature extraction capability. As shown in Tab.~\ref{tab:rank}, we observe that the model achieves the best performance when $r=64$.

\subsection{Analysis on Grouping}
We fix the number of points per group, resulting in dynamic group counts across different scenes. As shown in Tab.~\ref{tab:grouping}, using a fixed number of points yields performance slightly inferior to our method. This may be because fixed-point sampling struggles to adapt to point clouds with varying numbers of points. Therefore, we ultimately adopt the strategy of using a fixed number of groups for grouping.

\subsection{Results on ScanNet200}
To further verify the robustness of our method, we perform experiments on the ScanNet200~\cite{rozenberszki2022language} dataset. Built upon the ScanNet~\cite{dai2017scannet} benchmark, ScanNet200 expands the labeled categories, making the segmentation task substantially more challenging. As shown in Tab.~\ref{tab:200}, our PointTPA obtains 83.0\% in allAcc, a 0.7\% improvement over DAPT~\cite{zhou2024dynamic}, and achieve 33.2\% in mIoU, surpassing PointGST~\cite{liang2025parameter} by 1.0\%.

\begin{table}[h]
    \centering
    \vspace{-5pt}
    \footnotesize
    \setlength{\tabcolsep}{1mm}
    \caption{Ablation on ranks and grouping strategies. Here, N denotes the number of points per group.}
    \vspace{-10pt}
    \begin{subtable}[t]{0.54\linewidth}
        \centering
        \caption{\centering Ablation on different ranks.}
        \begin{tabular}{@{}>{\centering\arraybackslash}p{0.9cm}*{3}{>{\centering\arraybackslash}p{0.9cm}}@{}}
            \toprule
            \multicolumn{1}{c}{$r$} & mIoU & mAcc & allAcc\\
            \midrule
            32 & 77.4 & 85.5 & 91.7\\
            \rowcolor{greenbg}64 & \textbf{78.4} & \textbf{86.3} & \textbf{92.2}\\
            96 & 77.9 & 85.9 & 92.2\\
            128 & 77.7 & 85.6 & 92.1\\
            \bottomrule
        \end{tabular}
        \label{tab:rank}
    \end{subtable}
    \hfill
    \begin{subtable}[t]{0.45\linewidth}
        \centering
        \caption{Ablation on grouping.}
        \begin{tabular}{@{}>{\centering\arraybackslash}p{0.9cm}*{2}{>{\centering\arraybackslash}p{0.9cm}}@{}}
            \toprule
            \multicolumn{1}{c}{N} & mIoU &mAcc \\
            \midrule
            100 & 78.1 &86.2\\
            200&77.9 &86.3\\
            400&77.8 &85.7\\
            \rowcolor{greenbg}\textbf{Ours} & \textbf{78.4} &\textbf{86.3}\\
            \bottomrule
        \end{tabular}
        \label{tab:grouping}
    \end{subtable}
    \vspace{-15pt}
\end{table}

\section{Detailed Time Discussion}
\begin{wrapfigure}{r}{0.15\textwidth}
    \vspace{-12pt} 
    \hspace{-28pt} 
    \centering
    \footnotesize
    \setlength{\tabcolsep}{2pt}
    \begin{tabular}{ccc}
        \hline
        Method & Train\tiny (ms) & Infer\tiny(ms)\\
        \hline
       SNG &+9 &+2  \\
        DPP &+10 &+0   \\
        \rowcolor{greenbg}
        Both &+22 &+2\\
        \hline
    \end{tabular}
    \vspace{-15pt} 
\end{wrapfigure}
 As shown in the table, DPP and SNG incur comparable training time per scene. During inference, they introduce negligible overhead compared to the setting without them, which is consistent with our expectations.

\section{More Analysis}
\begin{figure*}[h]
\vspace{5pt}
\centering
\includegraphics[width=\linewidth]{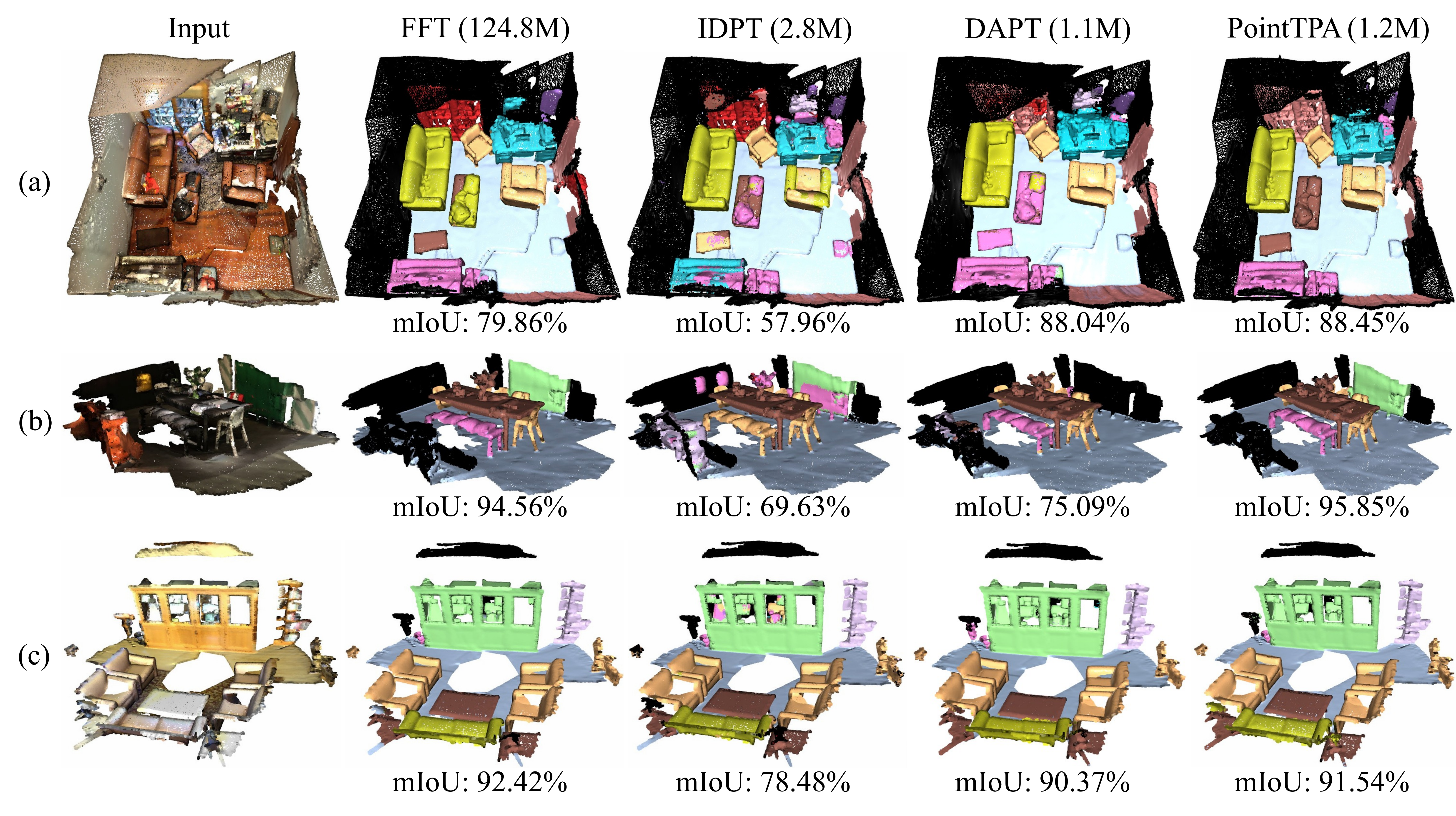}
\vspace{-10pt}
\caption{A comparison of FFT, IDPT~\cite{zha2023instance}, DAPT~\cite{zhou2024dynamic}, and PointTPA on segmentation performance, evaluated on ScanNet~\cite{dai2017scannet}.}
\label{comparison}
\end{figure*}

\paragraph{Segmentation Comparison}
Fig.~\ref{comparison} provides qualitative comparisons between FFT, 3D PEFT methods (\eg, IDPT~\cite{zha2023instance}, DAPT~\cite{zhou2024dynamic}), and our PointTPA. As highlighted in Fig.~\ref{comparison}(a), PointTPA uniquely distinguishes the tables from the sofa where other methods fail. Furthermore, in Fig.~\ref{comparison}(b), the predictions of PointTPA for fine-grained structures like bench legs are significantly closer to the FFT than IDPT and DAPT. These results demonstrate PointTPA's superior local feature extraction and its robustness in adapting to complex 3D scenes.

\begin{table}[t]
  \caption{Validation results on ScanNet200~\cite{rozenberszki2022language}.}
  \footnotesize
  \vspace{-10pt}
  \centering
  \resizebox{\columnwidth}{!}{%
  \begin{tabular}{@{}l*{6}{c}@{}}
    \toprule
    \multirow{2.3}{*}{ } & \multicolumn{2}{c}{\multirow{2.3}{*}{Param. (M)}}&\multirow{2.3}{*}{Reference } & \multicolumn{2}{c}
    {ScanNet200 Val~\cite{rozenberszki2022language}}  \\
     \cmidrule(lr){5-6} 
     & &  & &allAcc&mIoU \\
    \midrule
    \multicolumn{6}{l}{\textit{Full Fine-tuning}}\\
    \midrule
    Sonata(\text{w/o dec.})~\cite{wu2025sonata}  & \multicolumn{2}{c}{108.5}  &CVPR 25 & 83.8 &33.9   \\
    \midrule
    \multicolumn{6}{l}{\textit{General PEFT methods}}\\
    \midrule
    Sonata(\textit{linear probing}) &\multicolumn{2}{c}{0.2 (0.2\%)}  &CVPR 25 &81.1 &28.4 \\
    + Prefix Tuning~\cite{li2021prefix} &\multicolumn{2}{c}{1.2 (1.1\%)}  &ACL 21 &81.5\dplus{+0.4} &29.4\dplus{+1.0} \\
    + BitFit~\cite{zaken2022bitfit}  &\multicolumn{2}{c}{0.4 (0.4\%)}  &ACL 22 &82.3\dplus{+1.2} &31.1\dplus{+2.7}\\
    + LoRA~\cite{hu2021lora} &\multicolumn{2}{c}{1.2 (1.1\%)}   &ICLR 22 &\underline{82.5}\dplus{+1.4} &31.7\dplus{+3.3} \\
    + VeRA~\cite{kopiczkovera} &\multicolumn{2}{c}{0.3 (0.3\%)} &ICLR 24 &81.8\dplus{+0.7} &29.8\dplus{+1.4} \\
    + RandLoRA~\cite{albert2024randlora} &\multicolumn{2}{c}{0.9 (0.8\%)} &ICLR 25 &82.0\dplus{+0.9}  &30.3\dplus{+1.9} \\
    \midrule
    \multicolumn{6}{l}{\textit{PEFT methods for point cloud}}\\
    \midrule
    + IDPT~\cite{zha2023instance} &\multicolumn{2}{c}{2.8 (2.6\%)}  &ICCV 23 &81.2\dplus{+0.1} &27.9\dtplus{-0.5}\\
    + DAPT~\cite{zhou2024dynamic} &\multicolumn{2}{c}{1.4 (1.3\%)}  &CVPR 24 &82.3\dplus{+1.2} &\underline{33.2}\dplus{+4.8}\\
    + PointGST~\cite{liang2025parameter} &\multicolumn{2}{c}{1.3 (1.2\%)} &TPAMI 25 &82.1\dplus{+1.0} &32.2\dplus{+3.8}\\
    \midrule
    + PointTPA (\text{ours}) &\multicolumn{2}{c}{1.4 (1.3\%)}  &-- &\textbf{83.0}\dplus{+1.9} &\textbf{33.2}\dplus{+4.8}\\
    \bottomrule
  \end{tabular}
  }
  \vspace{-10pt}
  \label{tab:200}
\end{table}

\begin{figure*}[t]
\centering
\includegraphics[width=\linewidth]{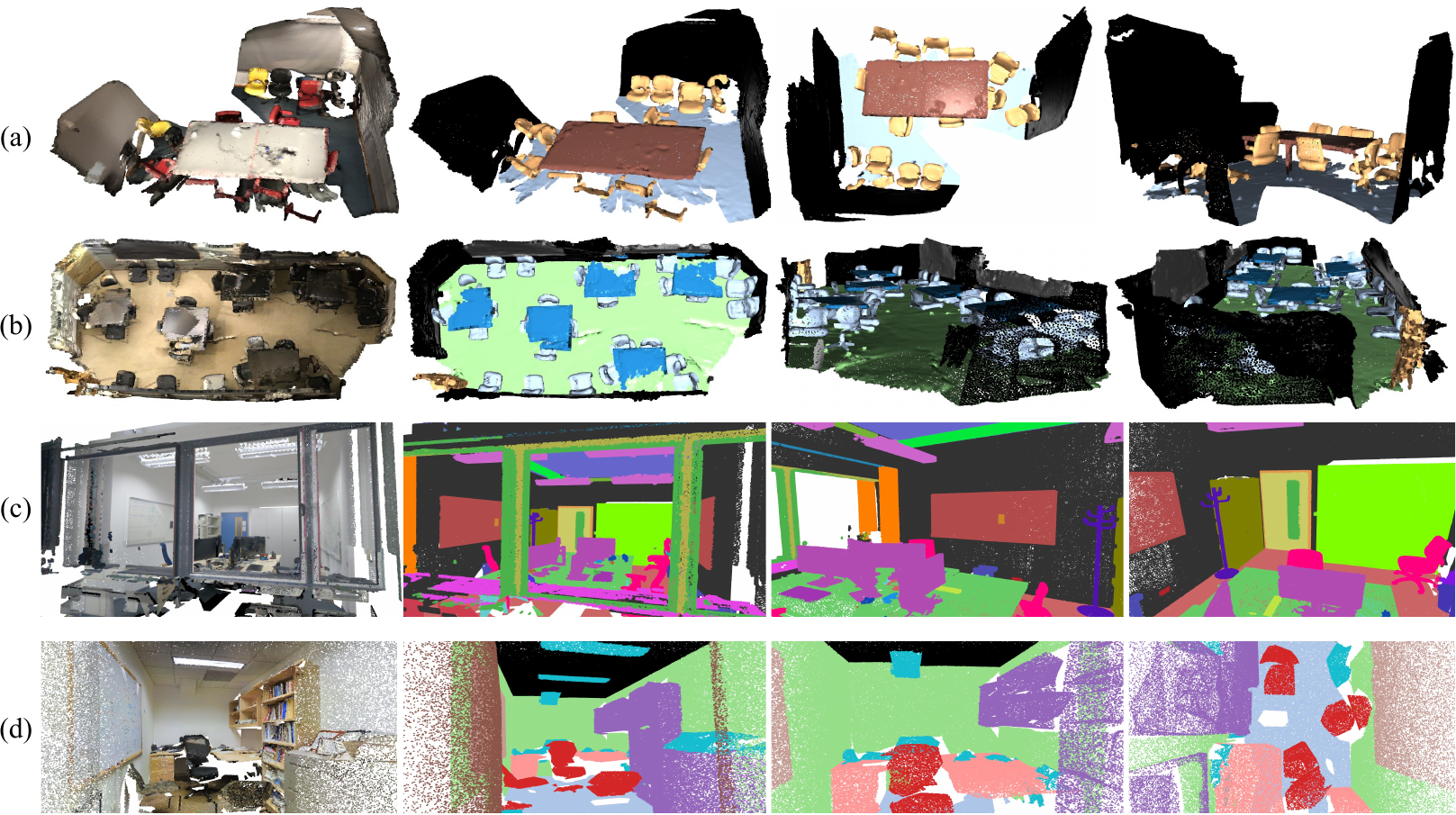}
\vspace{0pt}
\caption{More visualizations of the semantic segmentation results of our PointTPA on four large-scale scene datasets. (a) ScanNet~\cite{dai2017scannet}, (b) ScanNet200~\cite{rozenberszki2022language}, (c) ScanNet++~\cite{yeshwanth2023scannet++}, (d) S3DIS~\cite{armeni20163d} with 3 views.}
\label{vis}
\vspace{-15pt}
\end{figure*}

\paragraph{Qualitative Analysis}
In Fig.~\ref{vis}, we provide additional visualization results on ScanNet, ScanNet200, ScanNet++ and S3DIS with different viewpoints. In detail, we select one scenario with different viewpoints from the original scenario point cloud of each dataset. It is evident that PointTPA produces superior segmentation results.

\section{Test on ScanNet}
As shown in Tab.~\ref{tab:Test}, we evaluate our method on the ScanNet test set. PointTPA outperforms DAPT by 1.7\% and achieves performance on par with the FFT reference, with a marginal gap of 0.1\%. These results further demonstrate the strong generalizability of PointTPA.
 \begin{table}[h]
        \vspace{-5pt}
        \centering
        \footnotesize
        \setlength{\tabcolsep}{3.3mm} 
        \caption{Test on ScanNet~\cite{dai2017scannet}}
        \vspace{-5pt}
        \begin{tabular}{@{}>{}p{2.5cm}*{2}{>{\centering\arraybackslash}p{1.2cm}}@{}}
            \toprule
             Method & Reference  & mIoU\\
            \midrule
            Sonata (w/o dec.)&CVPR 25 &75.9\\
            \midrule
            Sonata (lin)&CVPR 25  &70.0\\
            + DAPT &CVPR 24 &74.1\dplus{+4.1}\\
            \rowcolor{greenbg}
            \multicolumn{1}{l@{\hskip -3pt}}{\hspace{-8.7pt}{+ \textbf{PointTPA (ours)}}} & --  &\multicolumn{1}{c@{\hskip 10pt}}{\hspace{3pt}{\textbf{75.8}\dplus{+5.8}}}  \\

            \bottomrule
        \end{tabular}
        \label{tab:Test}
    \end{table}

\end{document}